\documentclass[acmtog, authorversion]{acmart}

\newlength\paramargin
\newlength\figmargin

\newlength\secmargin
\newlength\figcapmargin
\newlength\tabcapmargin

\newcommand{\mpage}[2]
{
\begin{minipage}{#1\linewidth}\centering
#2
\end{minipage}
}

\newcommand{\topic}[1]
{
\vspace{1mm}\noindent\textbf{#1}
}

\newcommand{\figref}[1]{Figure~\ref{fig:#1}}

\long\def\ignorethis#1{}

\newbox\jsavebox%

\def\xi{\mathbf{x}_i}

\graphicspath{{figure}, {example}}

\usepackage{booktabs} %

\usepackage[ruled]{algorithm2e} %

\SetAlFnt{\small}
\SetAlCapFnt{\small}
\SetAlCapNameFnt{\small}
\SetAlCapHSkip{0pt}

\setcopyright{acmlicensed}
\acmJournal{TOG}
\acmYear{2021}
\acmVolume{40}
\acmNumber{6}
\acmArticle{218}
\acmMonth{12} 
\acmDOI{10.1145/3478513.3480559}

\begin{document}
\title{Pose with Style: Detail-Preserving Pose-Guided Image Synthesis with Conditional StyleGAN}

\author{Badour AlBahar}
\affiliation{%
  \institution{Virginia Tech},
  \institution{Kuwait University}
  }
\email{badour@vt.edu}
\author{Jingwan Lu}
\affiliation{%
 \institution{Adobe Research}
}
\email{jlu@adobe.com}
\author{Jimei Yang}
\affiliation{%
 \institution{Adobe Research}
}
\email{jimyang@adobe.com}
\author{Zhixin Shu}
\affiliation{%
  \institution{Adobe Research}
}
\email{zshu@adobe.com}
\author{Eli Shechtman}
\affiliation{%
  \institution{Adobe Research}
}
\email{elishe@adobe.com}
\author{Jia-Bin Huang}
\affiliation{%
  \institution{Virginia Tech},
 \institution{University of Maryland College Park}
}
\email{jbhuang@vt.edu}

\makeatletter
\let\@authorsaddresses\@empty
\makeatother

\begin{abstract}
We present an algorithm for re-rendering a person from a single image under arbitrary poses.
Existing methods often have difficulties in hallucinating occluded contents photo-realistically while preserving the identity and fine details in the source image.
We first learn to inpaint the correspondence field between the body surface texture and the source image with a human body symmetry prior.
The inpainted correspondence field allows us to transfer/warp local features extracted from the source to the target view even under large pose changes.
Directly mapping the warped local features to an RGB image using a simple CNN decoder often leads to visible artifacts.
Thus, we extend the StyleGAN generator so that it takes pose as input (for controlling poses) and introduces a spatially varying modulation for the latent space using the warped local features (for controlling appearances).
We show that our method 
compares favorably against the state-of-the-art algorithms in both quantitative evaluation and visual comparison. 
\end{abstract}



\begin{teaserfigure}
\newlength\figwidthA
\setlength\figwidthA{0.6\linewidth}

\newlength\figwidthB
\setlength\figwidthB{0.3\linewidth}

\begin{center}
\centering
\vspace{-3mm}

\mpage{0.03}{\raisebox{0pt}{\rotatebox{90}{Source image/Target pose}}}  \hfill
\hspace{-4mm}
\mpage{0.14}{\includegraphics[width=0.95\linewidth, trim=60 0 32 0, clip]{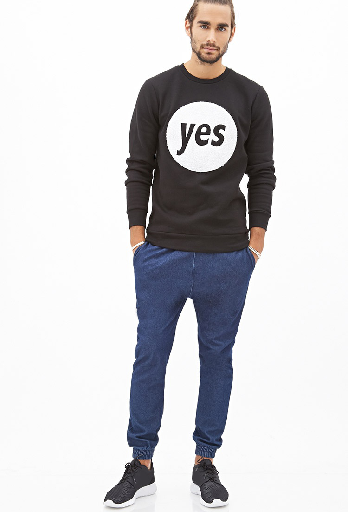}\llap{\includegraphics[height=1.75cm]{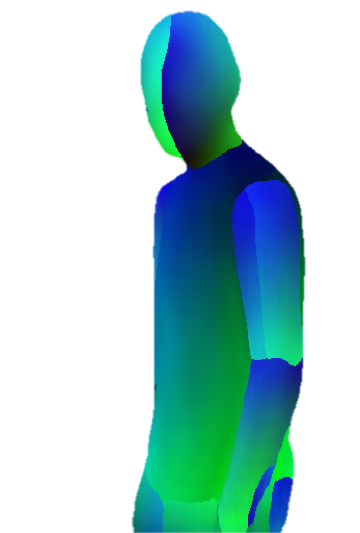}}}\hfill
\mpage{0.14}{\includegraphics[width=0.95\linewidth, trim=60 0 32 0, clip]{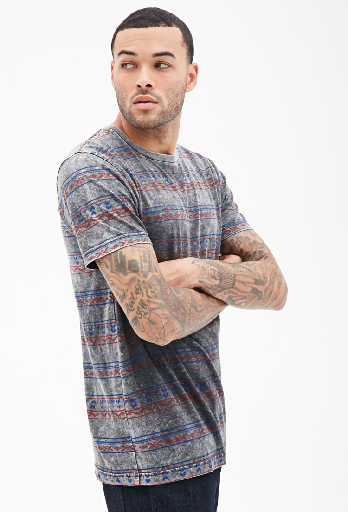}\llap{\includegraphics[height=1.75cm]{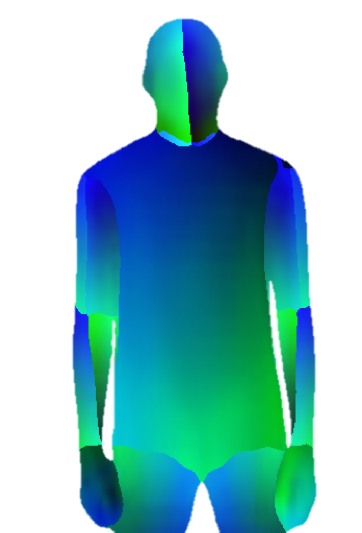}}}\hfill
\mpage{0.14}{\includegraphics[width=0.95\linewidth, trim=60 0 32 0, clip]{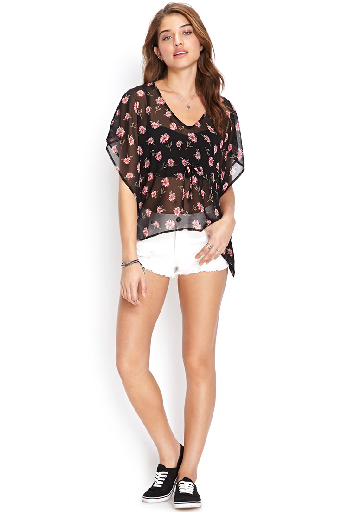}\llap{\includegraphics[height=1.75cm]{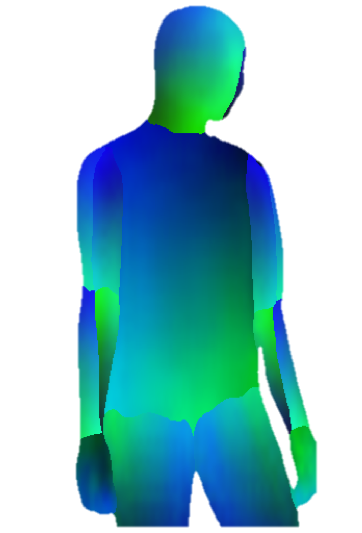}}}\hfill
\hspace{2.5mm}
\mpage{0.14}{\includegraphics[width=0.95\linewidth, trim=60 0 32 0, clip]{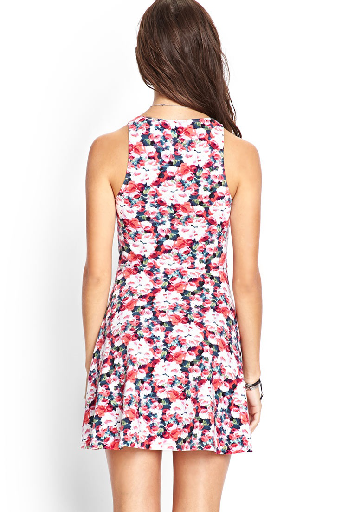}\llap{\includegraphics[height=1.75cm]{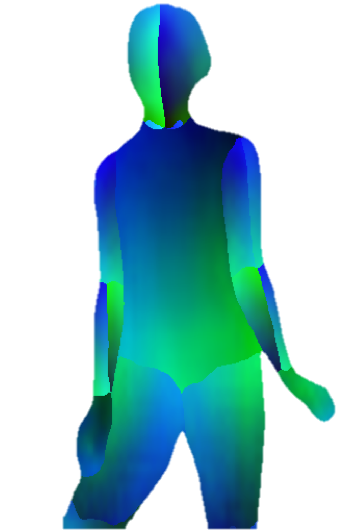}}}\hfill
\hspace{1.5mm}
\mpage{0.03}{\raisebox{0pt}{\rotatebox{90}{Source image/Target garment}}}  \hfill
\hspace{-4mm}
\mpage{0.14}{\includegraphics[width=0.95\linewidth, trim=60 0 32 0, clip]{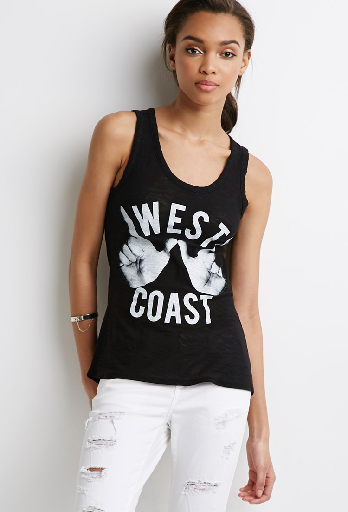}\llap{\includegraphics[height=2.5cm]{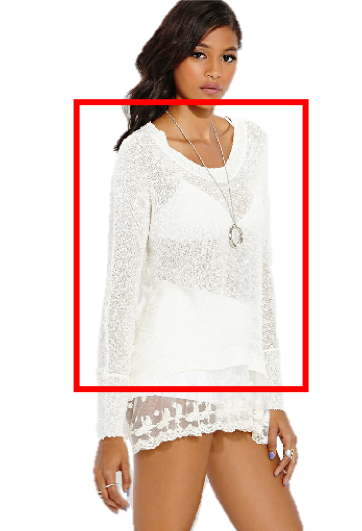}}}\hfill
\hspace{-4mm}
\mpage{0.14}{\includegraphics[width=0.95\linewidth, trim=60 0 32 0, clip]{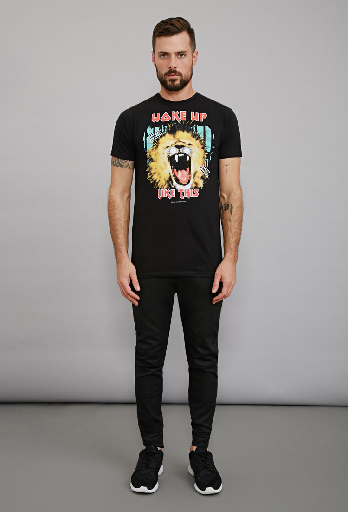}\llap{\includegraphics[height=2.5cm]{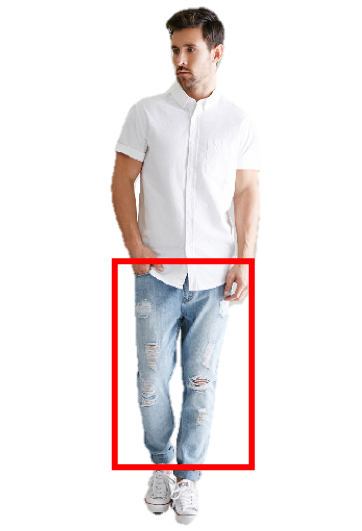}}}\\
\vspace{-1mm}
\mpage{0.03}{\raisebox{0pt}{\rotatebox{90}{Our results}}}  \hfill
\hspace{-12pt}
\mpage{0.14}{\includegraphics[width=0.95\linewidth, trim=60 0 32 0, clip]{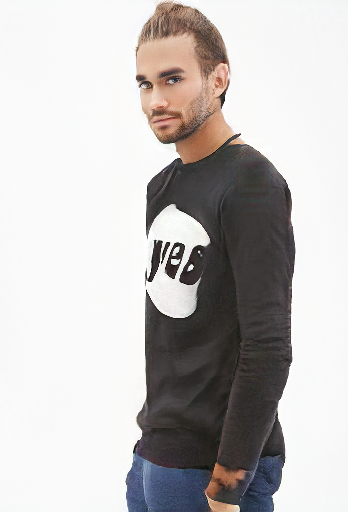}}\hfill
\mpage{0.14}{\includegraphics[width=0.95\linewidth, trim=60 0 32 0, clip]{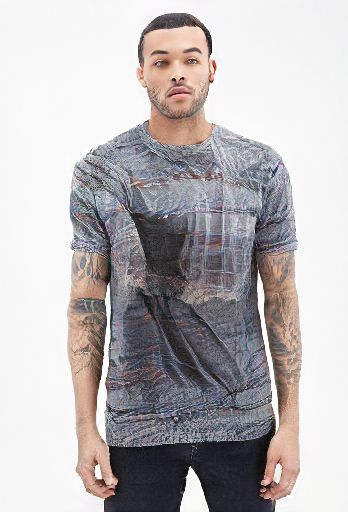}}\hfill
\mpage{0.14}{\includegraphics[width=0.95\linewidth, trim=60 0 32 0, clip]{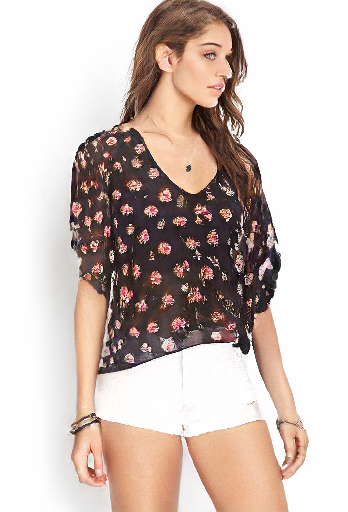}}\hfill
\hspace{4pt}
\mpage{0.14}{\includegraphics[width=0.95\linewidth, trim=60 0 32 0, clip]{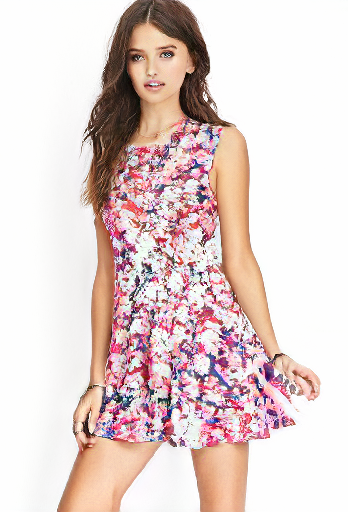}}\hfill
\hspace{3pt}
\mpage{0.03}{\raisebox{0pt}{\rotatebox{90}{Our results}}}  \hfill
\hspace{-10pt}
\mpage{0.14}{\includegraphics[width=0.95\linewidth, trim=60 0 32 0, clip]{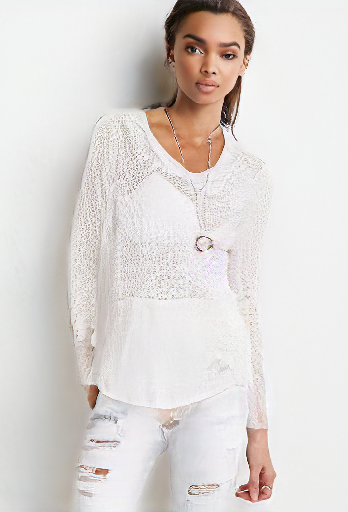}}\hfill
\hspace{-4mm}
\mpage{0.14}{\includegraphics[width=0.95\linewidth, trim=60 0 32 0, clip]{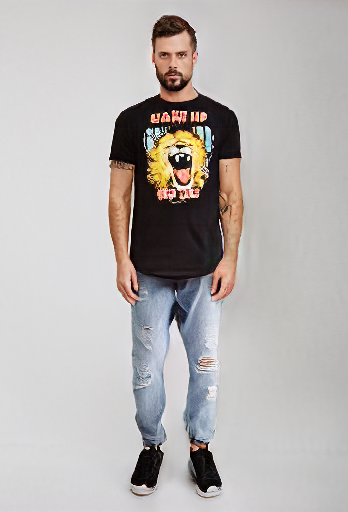}}\\

\hspace{8mm}
\mpage{0.62}{$\underbrace{\hspace{\textwidth}}_{\substack{\vspace{-7.0mm}\\\colorbox{white}{~~Pose-guided image synthesis~~}}}$}\hfill
\mpage{0.3}{$\underbrace{\hspace{\textwidth}}_{\substack{\vspace{-7.0mm}\\\colorbox{white}{~~Virtual try-on~~}}}$}\\

\vspace{-5mm}
\captionof{figure}{
\textbf{Detail-preserving pose-guided person image generation.}
We present a single-image human reposing algorithm guided by arbitrary body shapes and poses.
Our method first transfers local appearance features in the source image to the target pose with a human body symmetry prior.  
We then leverage a pose-conditioned StyleGAN2 generator with spatial modulation to produce photo-realistic reposing results.
Our work enables applications of posed-guided synthesis (\emph{left}) and virtual try-on (\emph{right}). Thanks to spatial modulation, our result preserves the texture details of the source image better than prior work. %
}
\label{fig:teaser}
\end{center}
\end{teaserfigure}

\maketitle

\section{Introduction}
\label{sec:intro}

Controllable, photo-realistic human image synthesis has a wide range of applications, including virtual avatar creation, reposing, virtual try-on, motion transfer, and view synthesis. 
Photo-realistic rendering of human images is particularly challenging through traditional computer graphics pipelines because it involves 1) designing or capturing 3D geometry and appearance of human and garments, 2) controlling poses via skeleton-driven deformation of 3D shape, and 3) synthesizing complicated wrinkle patterns for loose clothing.
Recent learning-based approaches alleviate these challenges and have shown promising results. 
These methods typically take inputs 1) a single source image capturing the human appearance and 2) a target pose representation (part confidence maps, skeleton, mesh, or dense UV coordinates) 
and synthesize a novel human image with the appearance from source and the pose from the target.

\emph{Image-to-image translation} based methods~\cite{ma2017pose,ma2018disentangled,pumarola2018unsupervised,esser2018variational,siarohin2018deformable}, building upon conditional generative adversarial networks~\cite{isola2017image}, learn to predict the reposed image from the source image and the target pose.
However, as human reposing involves significant spatial transformations of appearances, such approaches often require per-subject training using multiple images from the same persons~\cite{chan2019everybody,wang2018video} or are incapable of preserving the person's identity and the fine appearance details of the clothing in the source image.

\emph{Surface-based} approaches~\cite{neverova2018dense,grigorev2019coordinate,lazova2019360,sarkar2020neural} map human pixels in the source image to the canonical 3D surface of the human body (e.g., SMPL model~\cite{loper2015smpl}) with part segmentation and UV parameterization. 
This allows transferring pixel values (or local features) of visible human surfaces in the input image to the corresponding spatial location specified by the target pose. 
These methods thus retain finer-grained local details and identity compared to image-to-image translation models.
However, modeling human appearance as a single UV texture map cannot capture view/pose-dependent appearance variations and loose clothing. 

\emph{StyleGAN-based} methods~\cite{lewis2020vogue,ADGAN_2020,sarkar2021style} very recently have shown impressive results for controllable human image synthesis~\cite{ADGAN_2020,sarkar2021style} or virtual try-on \cite{lewis2020vogue}. 
The key ingredient is to extend the unconditioned StyleGAN network~\cite{Karras2019stylegan2} to a \emph{pose-conditioned} one. 
While their generated images are photo-realistic, it remains challenging to preserve fine appearance details (e.g., unique patterns/textures of garments) in the source image due to the global (spatially-invariant) modulation/demodulation of latent space.

\begin{figure}[t]
\centering
\vspace{-4mm}
\mpage{0.03}{\raisebox{0pt}{\rotatebox{90}{\small{PATN~\cite{PATN_2019}}}}}  \hfill
\hspace{-3mm}
\mpage{0.22}{\includegraphics[width=0.95\linewidth, trim=20 0 20 0, clip]{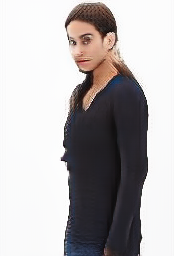}}\hfill
\mpage{0.22}{\includegraphics[width=0.95\linewidth, trim=20 0 20 0, clip]{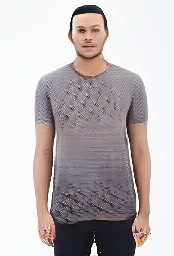}}\hfill
\mpage{0.22}{\includegraphics[width=0.95\linewidth, trim=20 0 20 0, clip]{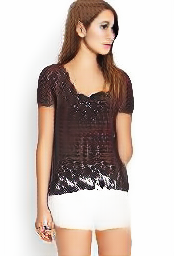}}\hfill
\mpage{0.22}{\includegraphics[width=0.95\linewidth, trim=20 0 20 0, clip]{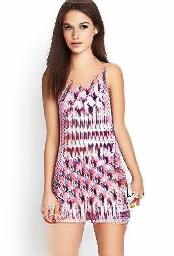}}\\
\vspace{-1mm}
\mpage{0.03}{\raisebox{0pt}{\rotatebox{90}{\small{ADGAN~\cite{ADGAN_2020}}}}}  \hfill
\hspace{-3mm}
\mpage{0.22}{\includegraphics[width=0.95\linewidth, trim=20 0 20 0, clip]{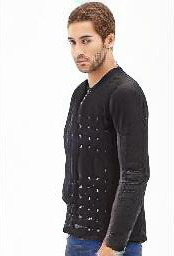}}\hfill
\mpage{0.22}{\includegraphics[width=0.95\linewidth, trim=20 0 20 0, clip]{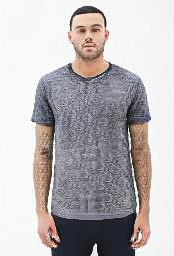}}\hfill
\mpage{0.22}{\includegraphics[width=0.95\linewidth, trim=20 0 20 0, clip]{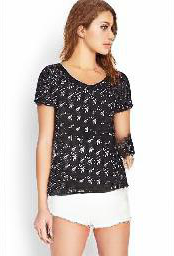}}\hfill
\mpage{0.22}{\includegraphics[width=0.95\linewidth, trim=20 0 20 0, clip]{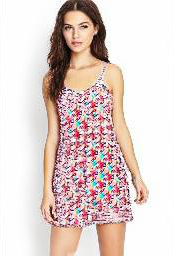}}\\
\vspace{-1mm}
\mpage{0.03}{\raisebox{0pt}{\rotatebox{90}{\small{GFLA~\cite{GFLA_2020}}}}}  \hfill
\hspace{-3mm}
\mpage{0.22}{\includegraphics[width=0.95\linewidth, trim=40 0 40 0, clip]{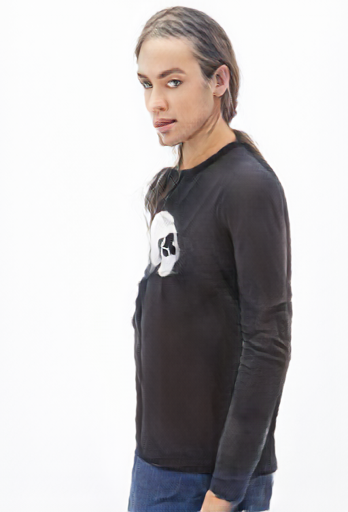}}\hfill
\mpage{0.22}{\includegraphics[width=0.95\linewidth, trim=40 0 40 0, clip]{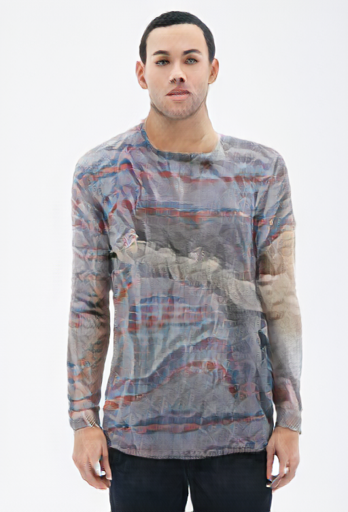}}\hfill
\mpage{0.22}{\includegraphics[width=0.95\linewidth, trim=40 0 40 0, clip]{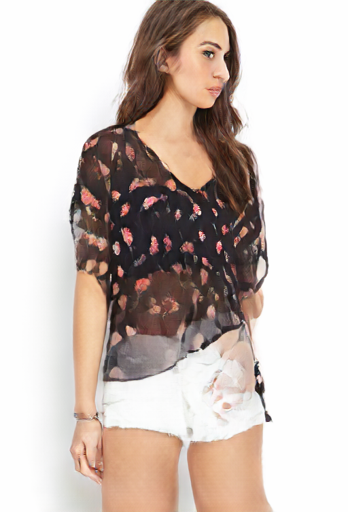}}\hfill
\mpage{0.22}{\includegraphics[width=0.95\linewidth, trim=40 0 40 0, clip]{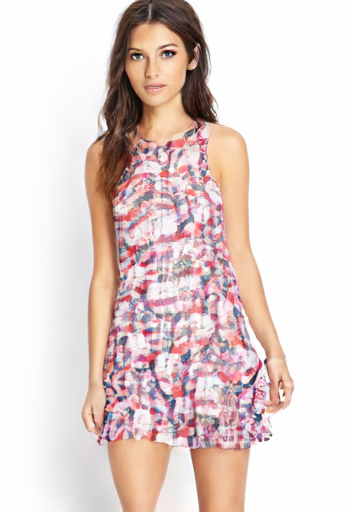}}\\
\vspace{-3mm}
\caption{\textbf{Limitations of existing methods.}
Existing human reposing methods struggle to preserve details in the source image. 
Common issues include identity (1{st} and 2{nd} columns) and clothing textures (3{rd}, 4{th} columns) changes. 
Compare these results with ours in \figref{teaser}.
}
\vspace{-3mm}
\label{fig:limitation}
\end{figure}

We present a new algorithm for generating \emph{detail-preserving} and 
\emph{photo-realistic} re-rendering of human with novel poses from a \emph{single source image}. 
Similar to the concurrent work~\cite{sarkar2021style,lewis2020vogue}, we use a pose-conditioned StyleGAN network for generating pose-guided images.
To preserve fine-grained details in the source image, we learn to inpaint the correspondence field between 3D body surface and the source image using a body symmetry prior.
Using this inpainted correspondence field, we transfer local features from the source to the target pose and use the warped local features to modulate the StyleGAN generator network at multiple StyleBlocks in a \emph{spatially varying} manner.
As we combine complementary techniques of photo-realistic image synthesis (from StyleGAN-based methods) and the 3D-aware detail transfers (from surface-based methods), our method achieves high-quality human-reposing and garment transfer results (\figref{teaser}) and alleviates visible artifacts compared with the state-of-the-art (\figref{limitation}). While StylePoseGan~\cite{sarkar2021style} (concurrent work to ours) also combines pose-conditioned StyleGAN with the use of proxy geometry, its global modulation/demodulation scheme limits its ability to preserve fine appearance details. We evaluate the proposed algorithm visually and quantitatively using the DeepFashion dataset~\cite{liuLQWTcvpr16DeepFashion} and show favorable results compared to the current best-performing methods. \emph{Our contributions} include:
\begin{itemize}
\item We integrate the techniques from surface-based and styleGAN-based methods to produce \emph{detail-preserving} and \emph{photo-realistic} controllable human image synthesis.
\item We propose an explicit \emph{symmetry prior} of the human body for learning to inpaint the correspondence field between human body surface and the source image which facilitates detail transfer, particularly for drastic pose changes.
\item We present a \emph{spatially varying} variant of latent space modulation in the StyleGAN network, allowing us to transfer local details while preserving photo-realism.
\end{itemize}

\section{Related Work}
\label{sec:related}

\topic{Pose-guided Person Image Synthesis}
aims to transfer a person's appearance from a source image to a specified target pose.
Example applications include motion transfer~\cite{chan2019everybody,aberman2019deep,yoon2020pose}, human reposing~\cite{ma2017pose,ma2018disentangled,ADGAN_2020}, and virtual try-on~\cite{ADGAN_2020,lewis2020vogue,sarkar2021style}.
These approaches typically encode the poses in the source and target views as either part confidence maps~\cite{aberman2019deep,ma2017pose,ma2018disentangled} or  skeleton~\cite{chan2019everybody,pumarola2018unsupervised,esser2018variational,siarohin2018deformable,PATN_2019,ADGAN_2020,GFLA_2020} 
and use a conditional generative adversarial network to produce the reposed images.
To handle large pose changes between the source and target views, existing methods leverage per-subject training~\cite{chan2019everybody,wang2018video,aberman2019deep}, spatial transformation/deformation~\cite{siarohin2018deformable,balakrishnan2018synthesizing,GFLA_2020}, and local attention~\cite{GFLA_2020}. To better retain the person identity and detailed appearance in the source image, surface-based methods first establish the correspondence between pixels from the source/target image to a canonical coordinate system of the 3D human body (with UV parameterization).
These methods can then transfer pixel values~\cite{neverova2018dense,grigorev2019coordinate,alldieck2019tex2shape,lazova2019360} or local features~\cite{sarkar2020neural} to the target pose.
As the commonly used UV parameterization only captures the surface of a tight human body~\cite{loper2015smpl}
(i.e., cannot encode loose clothing), several methods alleviate this through explicitly predicting garment labels at the target view~\cite{yoon2020pose} or implicitly re-rendering the warped features~\cite{sarkar2020neural}.
Very recently, pose-conditioned StyleGAN networks have been proposed~\cite{ADGAN_2020,lewis2020vogue,sarkar2021style}.
Such methods control the target appearance using the appearance features extracted from the source image~\cite{ADGAN_2020,lewis2020vogue} or pose-independent UV texture~\cite{sarkar2021style} to modulate the latent space. 

Our method builds upon pose-conditioned StyleGAN but differs from prior work in two critical aspects.
First, instead of \emph{global} latent feature modulation used in prior work, we propose to use a \emph{spatially varying} modulation for improved local detail transfer. 
Second, we train a coordinate inpainting network for completing partial correspondence field (between the body surface and source image) using a human body symmetry prior. 
This allows us to directly transfer local features extracted from the source to the target pose.

\topic{Neural Rendering}
methods first render a coarse RGB image or neural textures via 
some forms of geometry proxy (e.g., point cloud, 2D/3D skeleton, or human surface meshes) and then map the coarse RGB or latent textures to an RGB image using a translation network~\cite{meshry2019neural,kim2018deep,thies2019deferred,raj2021anr,liu2020neural,tewari2020state}. 
A recent line of research focuses on learning volumetric neural scene representations for view synthesis~\cite{lombardi2019neural,mildenhall2020nerf}.
Such approaches have been extended to handle dynamic scenes (e.g., humans)~\cite{park2020deformable,tretschk2020non,xian2020space,li2020neural,gao2021dynamic}.
Recent efforts further focus on enabling controls over viewpoints~\cite{gao2020portrait,gafni2021dynamic}, pose~\cite{noguchi2021neural,peng2021animatable}, expressions~\cite{gafni2021dynamic}, illumination~\cite{zhang2021neural} of human face/body.
However, most of these approaches often require computationally expensive \emph{per-scene/per-person training} over multiple image observations.

Our method also uses body surface mesh as our geometry proxy for re-rendering. 
Instead of using a simple CNN translation network, we integrate the rendered latent texture with StyleGAN through spatially varying modulation.
In contrast to volumetric neural rendering techniques, our method does \emph{not} require per-subject training.

\topic{Deep Generative Adversarial Networks}
have shown great potentials for synthesizing high-quality photo-realistic images~\cite{goodfellow2014generative,zhang2019self,brock2018large,karras2017progressive,Karras2019stylegan2}.
Using the pre-trained model, several works discover directions in the latent space that correspond to spatial or semantic changes~\cite{shoshan2021gan,peebles2020hessian,harkonen2020ganspace,shen2021closed,jahanian2020steerability}. 
In the context of portrait images, some recent methods provide 3D control for the generated samples~\cite{tewari2020stylerig,abdal2021styleflow} or real photographs~\cite{tewari2020pie}.
Our work focuses on designing a pose-conditioned GAN with precise control on the localized appearance (for virtual try-on) and pose (for reposing).

\topic{Image-to-Image Translation}
provides a general framework for mapping an image from one visual domain to another~\cite{isola2017image,wang2017high,park2019SPADE}.
Recent advances include learning from unpaired dataset~\cite{zhu2017unpaired,huang2018multimodal,lee2018diverse}, extension to videos~\cite{wang2018video,wang2019few}, and talking heads~\cite{zakharov2019few,wang2021one}.

Similar to many existing human reposing methods~\cite{ma2017pose,ma2018disentangled,chan2019everybody,pumarola2018unsupervised,PATN_2019,ADGAN_2020,GFLA_2020}, our work can be cast an image-to-image translation problem that maps an input target pose to an RGB image with the appearance from a source image.
Our core technical novelties lie in 1) spatial modulation in StyleGAN for detail transfer and 2) a body symmetry prior for correspondence field inpainting.

\topic{Localized Manipulation} is often preferable over global changes in many image editing scenarios. 
Existing work addresses this via structured noise~\cite{alharbi2020disentangled}, local semantic latent vector discovery~\cite{chai2021using}, latent space regression~\cite{collins2020editing}, and explicit masking~\cite{shocher2020semantic}.

Our method for spatially varying feature modulation in StyleGAN shares high-level similarity with approaches that add spatial dimensions to the latent vectors in unconditional StyleGAN~\cite{alharbi2020disentangled,kim2021stylemapgan} and conditional GANs~\cite{park2019SPADE,albahar2019guided}.
Our work differs in that our spatial modulation parameters are predicted from the warped appearance features extracted from the source image instead of being generated from random noise using a mapping network.

\topic{Symmetry Prior}(in particular reflective symmetry) has been applied for learning deformable 3D objects~\cite{wu2020unsupervised}, 3D reconstruction of objects~\cite{sinha2012detecting}, and human pose regression~\cite{yeh2019chirality}.
Our work applies left-right reflective symmetry to facilitate the training of coordinate-based texture inpainting network. 
The symmetry prior allows us to better reuse local appearance features from the source and thus leads to improved results when source and target poses are drastically different.

\begin{figure*}[t]    
\centering
\includegraphics[width=\linewidth]{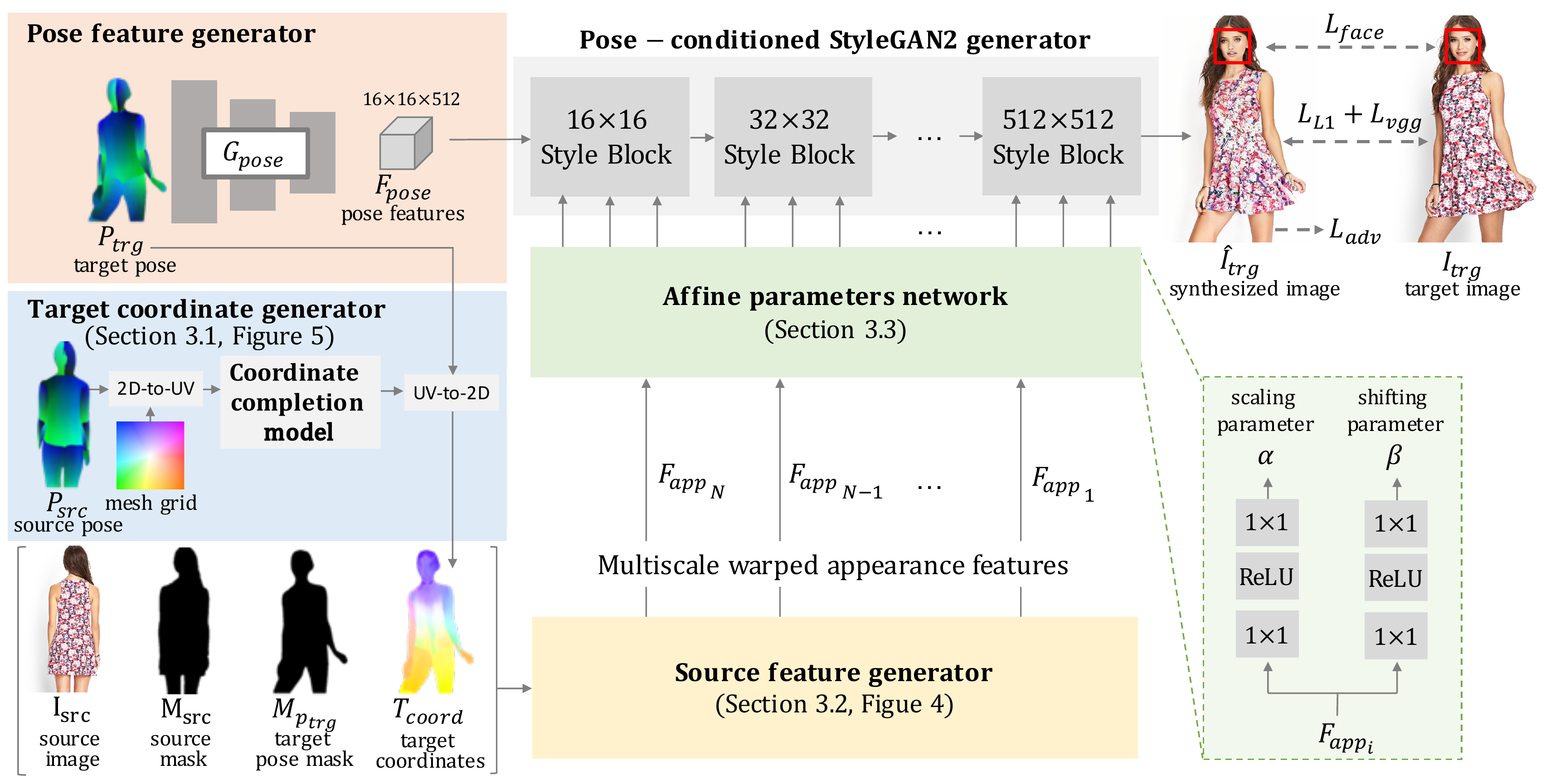}
\vspace{-6mm}
\caption{\textbf{Method overview.}
Our human reposing model builds upon a pose-conditioned StyleGAN2 generator~\cite{Karras2019stylegan2}. 
We extract the DensePose~\cite{guler2018densepose} representation $P_{trg}$ and use a pose encoder $G_{pose}$ to encode $P_{trg}$ into $16\times16\times512$ pose features $F_{pose}$ which is used as input to the StyleGAN2 generator~\cite{Karras2019stylegan2}.
To preserve the source image appearance, we encode the input source image $I_{src}$ into multiscale warped appearance features $F_{app_i}$ using the source feature generator (\figref{sfg}).
To warp the feature from the source pose to the target pose we use the target coordinates $T_{coord}$. We compute these target coordinates $T_{coord}$ using 1) the target dense pose $P_{trg}$ and 2) the completed coordinates in the UV-space inpainted using the coordinate completion model (\figref{coordinate_completion}). 
We pass the multi-scale warped appearance features $F_{app_i}$ through the affine parameters network to generate scaling and shifting parameters $\alpha$ and $\beta$ that are used to modulate the StyleGAN2 generator features in a \emph{spatially varying} manner (\figref{revised_modulation}). 
Our training losses include adversarial loss, reconstruction losses, and a face identity loss.
}
\label{fig:revised_overview}
\end{figure*}

\begin{figure}[t]
\centering
\includegraphics[width=\linewidth]{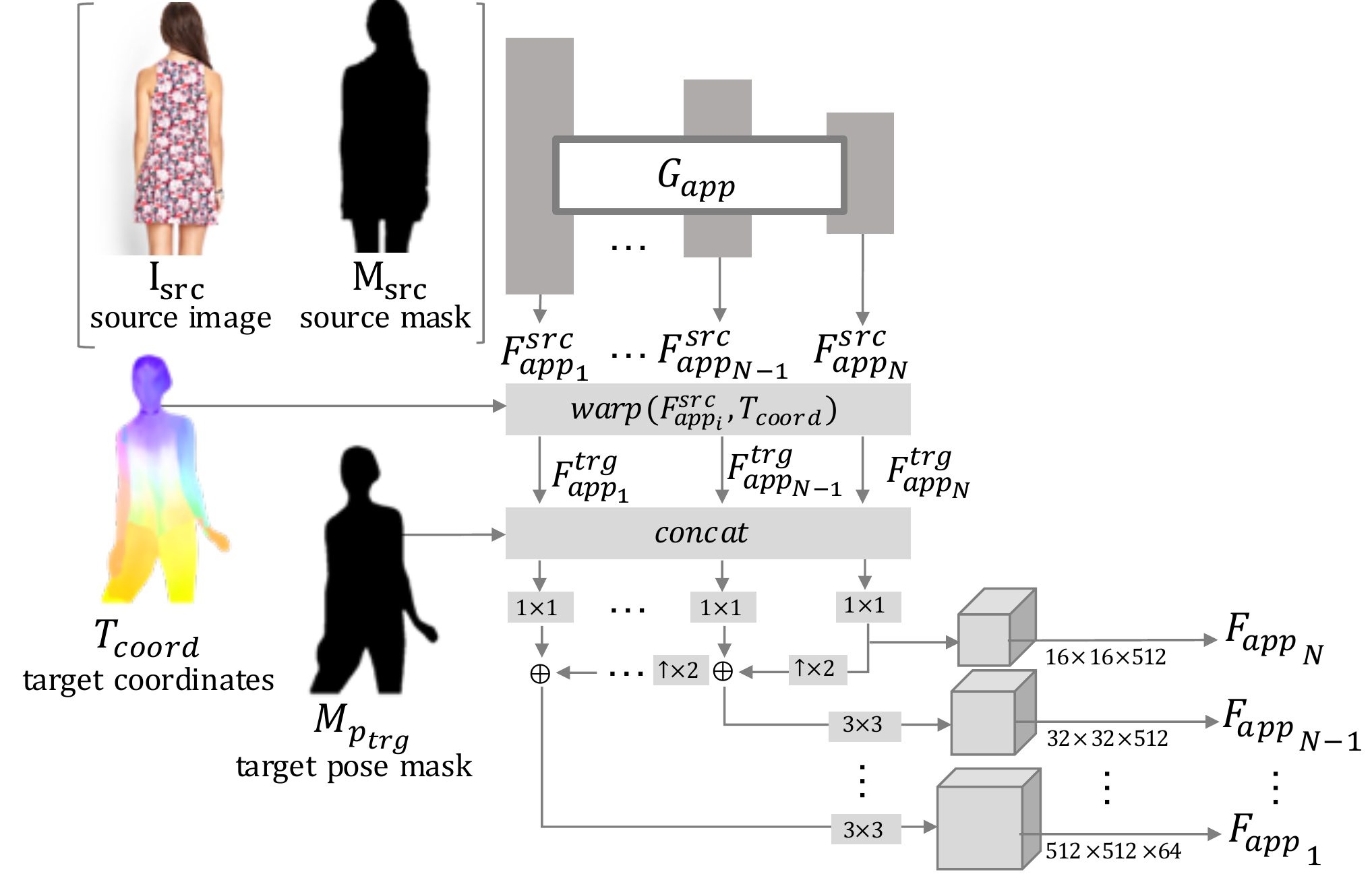}
\vspace{-4mm}
\caption{\textbf{Source feature generator.}
To preserve the source image appearance, we encode the input source image $I_{src}$ into multiscale features $F_{app_i}^{src}$ and warp them from the source pose to the target pose $F_{app_i}^{trg}$ using the target coordinates $T_{coord}$ computed from the target coordinate generator (\figref{revised_overview}). 
We further process the warped features with a feature pyramid network~\cite{lin2017feature} to obtain the multi-scale warped appearance features $F_{app_i}$ which go through affine parameters network to generate scaling and shifting parameters $\alpha$ and $\beta$ that are used to modulate the StyleGAN2 generator features in a \emph{spatially varying} manner (\figref{revised_modulation}). 
}
\label{fig:sfg}
\end{figure}

\section{Method}
\label{sec:method}

Given an image of a person $I_{src}$ and a desired target pose $P_{trg}$ 
represented by Image-space UV coordinate map per body part (shortly IUV) extracted from DensePose~\cite{guler2018densepose}, 
our goal is to generate an image preserving the appearance of the person in $I_{src}$ in the desired pose $P_{trg}$.
Note that this IUV representation of dense pose entangles both the pose and shape representation.

We show an overview of our proposed approach in \figref{revised_overview}.
We use a pose-guided StyleGAN2 generator~\cite{Karras2019stylegan2} that takes $16\times16\times512$ pose features $F_{pose}$ as input. 
The pose features $F_{pose}$ are encoded from the DensePose representation~\cite{guler2018densepose} using a pose feature generator $G_{pose}$ that is composed of several residual blocks.
Using the source and target pose, we use coordinate completion model to produce target coordinate that establishes the correspondence between target and source image.
To encode the appearance information, we use a feature pyramid network~\cite{lin2017feature} $G_{app}$ to encode the source image into multiscale features and warp them according to the target pose. 
We then use the warped appearance features to generate scaling and shifting parameters to spatially modulate the latent space of the StyleGAN generator.

\begin{figure*}[t]
\centering
\mpage{0.45}{\includegraphics[width=\linewidth]{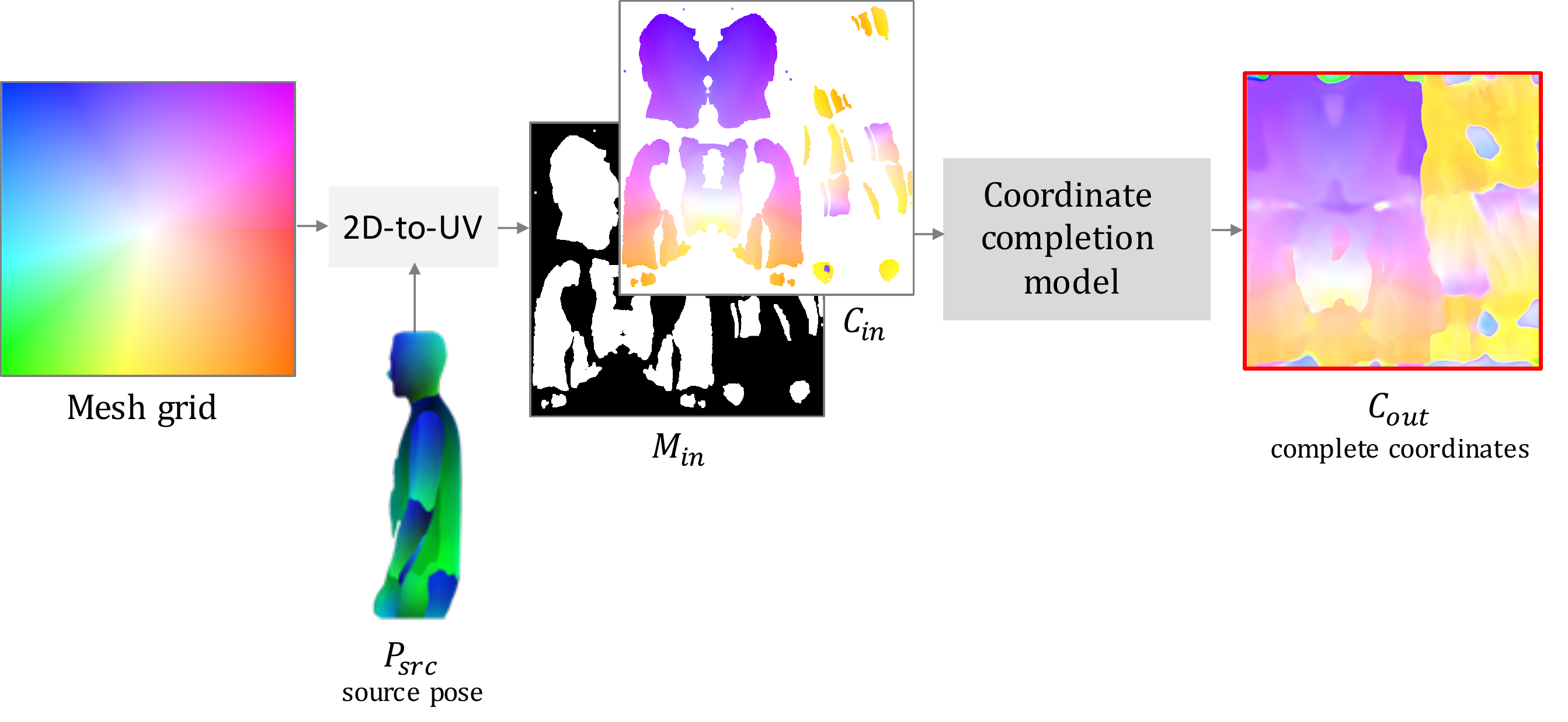}} \hfill
\mpage{0.52}{\includegraphics[width=\linewidth]{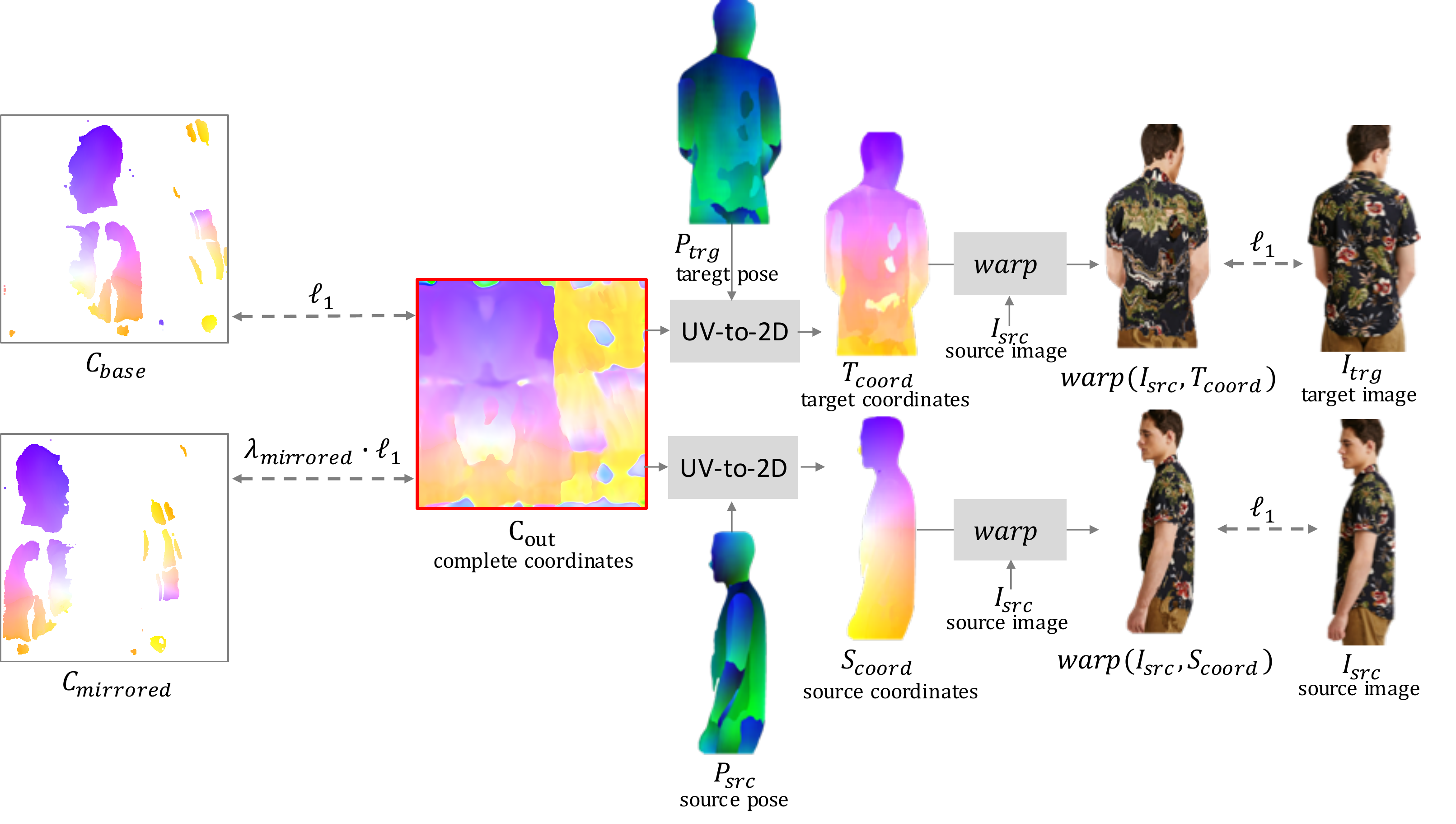}}
\\
\mpage{0.45}{\small{(a) Coordinate completion model}} \hfill
\mpage{0.52}{\small{(b) Training losses}}\\
\vspace{-2mm}
\caption{\textbf{Coordinate completion model.}
The goal of the coordinate completion model is to learn how to \emph{reuse} the local features of the visible parts of the human in the source image for the invisible parts (unseen in the source pose) in the target pose.
(a) Given a mesh grid and the dense pose of the input source image $P_{src}$, we map the base coordinates $C_{base}$ and their symmetric counterpart $C_{mirrored}$ from the 2D mesh grid to the UV-space using a pre-computed mapping table. 
We then concatenate the combined coordinates $C_{in}$ and their corresponding visibility mask $M_{in}$ as input to the coordinate completion model.
(b) We train the model to minimize the L1 loss between the predicted coordinates $C_{out}$ and the input coordinates $C_{in}$ as shown in Eqn.~\ref{eqn:Lcoor}. 
We also minimize the L1 loss between the warped source image and the warped target image as shown in Eqn.~\ref{eqn:Lrgb}. 
}

\label{fig:coordinate_completion}
\end{figure*}

\subsection{Coordinate Completion Model}
The IUV map of the source pose $P_{src}$ allows us to represent the \emph{pose-independent} appearance of the person in the UV-space. 
However, only the appearance of \emph{visible} body surface can be extracted.
This leads to incomplete UV-space appearance representation and thus may not handle the dis-occluded appearance for the target pose $P_{trg}$.
Previous work ~\cite{sarkar2021style} encodes the partial UV-space appearance to a \emph{global latent vector} for modulating the generator. 
This works well for clothing with uniform colors or homogeneous textures, but inevitably loses the spatially-distributed appearance details. 
We propose to inpaint the UV-space appearance by a neural network guided by the human body mirror-symmetry prior. 
Instead of directly inpainting pixel values in UV-space, we choose to complete the mapping from image-space to UV-space established by $P_{src}$ and represented by UV-space source image coordinates, in order to avoid generating unwanted appearance artifacts while best preserving the source appearance. We refer to this network as \emph{coordinate completion model}.
We show an overview of our coordinate completion model in Figure~\ref{fig:coordinate_completion}.

Given a mesh grid and the dense pose of the input source image $P_{src}$, we use a pre-computed image-space to/from UV-space mapping to map coordinates from the mesh grid to appropriate locations in the UV-space (using bilinear sampling for handling fractional coordinates). 
We denote these base mapped coordinates as $C_{base}$ and the mask indicating where these coordinates are as $M_{base}$. 

Since human appearances are often left-right symmetrical, in addition to these base coordinates, we also map the left-right mirrored coordinates to the UV-space $C_{mirrored}$ and denote their respective mask as $M_{mirrored}$. 
Visualization of mapped base and mirrored coordinates are shown in Figure~\ref{fig:symmetry}.

We combine the incomplete UV-space base and mirrored coordinates and their respective masks, such that:
\begin{equation}
M_{mirrored} = M_{mirrored} - (M_{base}\cdot M_{mirrored})
\end{equation}
\begin{equation}
M_{in} = M_{base} + M_{mirrored}
\end{equation}
\begin{equation}
C_{in} = C_{base}\cdot M_{base}+ C_{mirrored}\cdot M_{mirrored}
\end{equation}

We concatenate the combined coordinates $C_{in}$ and their mask $M_{in}$ and pass them as input to the coordinate completion model. To implement our coordinate completion model, we follow a similar architecture to the coordinate inpainting architecture proposed by~\cite{grigorev2019coordinate} with gated convolutions~\cite{yu2018free}. 

We train our model to minimize the $\ell_1$ loss between the generated coordinates $C_{out}$ and the input coordinates, such that:
\begin{equation}\label{eqn:Lcoor}
\begin{split}
L_{coord} = || C_{out} \cdot M_{base} - C_{base} \cdot M_{base} ||_{1} \\
+ \lambda_{mirrored} \cdot  ||C_{out} \cdot M_{mirrored} - C_{mirrored} \cdot M_{mirrored}||_{1},
\end{split}
\end{equation}
where $\lambda_{mirrored} $ is set to $0.5$.

We also utilize the source-target pairs to train the coordinate completion model. Specifically, we use the source dense pose $P_{src}$ to map the generated complete coordinates from the UV-space to the source image-space $S_{coord}$. Similarly, we also use the target dense pose $P_{trg}$ to map the generated complete coordinates from the UV-space to the target image-space $T_{coord}$ using the pre-computed mapping table.
We then use these target and source coordinates to warp pixels from the input source image $I_{src}$ and minimize the $\ell_1$ loss between the foreground of the warped images and the foreground of the ground truth images, such that:

\begin{equation}\label{eqn:Lrgb}
\begin{split}
L_{rgb} = || warp(I_{src}, S_{coord}) \cdot M_{P_{src}} - I_{src}\cdot M_{P_{src}} ||_{1}\\
+ || {warp(I_{src}, T_{coord}) \cdot M_{P_{trg}} - I_{trg}\cdot M_{P_{trg}}}||_{1},
\end{split}
\end{equation}
where $M_{P_{src}}$ and $M_{P_{trg}}$ are the source pose mask and target pose mask, respectively.

The total loss to train the coordinate completion model is:
\begin{equation}
L = L_{coord} + L_{rgb}
\end{equation}

\newcommand{\shiftleft}[2]{\makebox[0pt][r]{\makebox[#1][l]{#2}}}

\begin{figure*}[t]
\centering
\mpage{0.52}{
\mpage{0.03}{\raisebox{0pt}{\rotatebox{90}{\small{Coordinates/}}}}  \hfill
\hspace{-2mm}
\mpage{0.03}{\raisebox{0pt}{\rotatebox{90}{\small{Source pose}}}}  \hfill
\hspace{-3mm}
\mpage{0.3}{\fbox{\includegraphics[width=\linewidth]{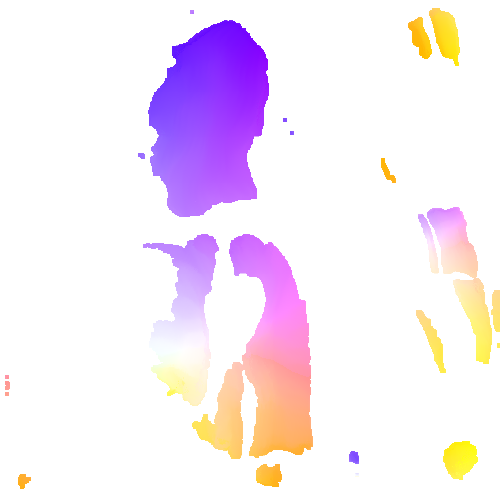}}\llap{\shiftleft{2.8cm}{\raisebox{1.6cm}{\includegraphics[height=1cm]{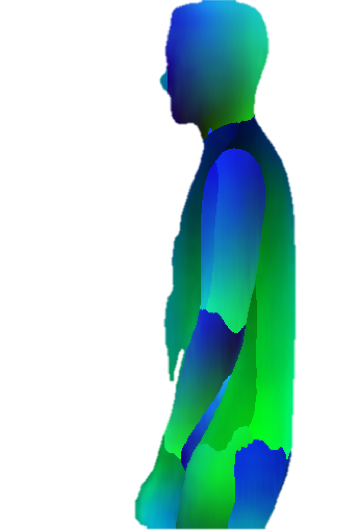}}}}}\hfill
\mpage{0.3}{\fbox{\includegraphics[width=\linewidth]{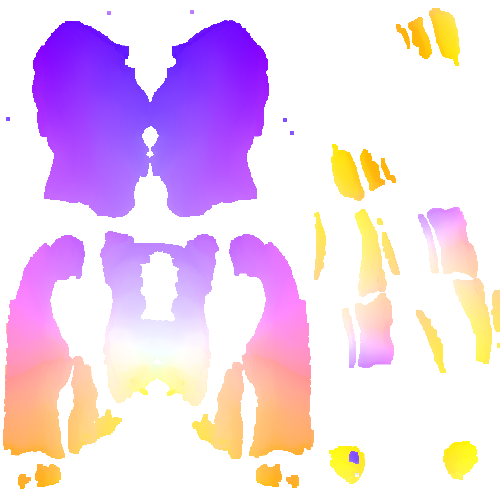}}}\hfill
\mpage{0.3}{\fbox{\includegraphics[width=\linewidth]{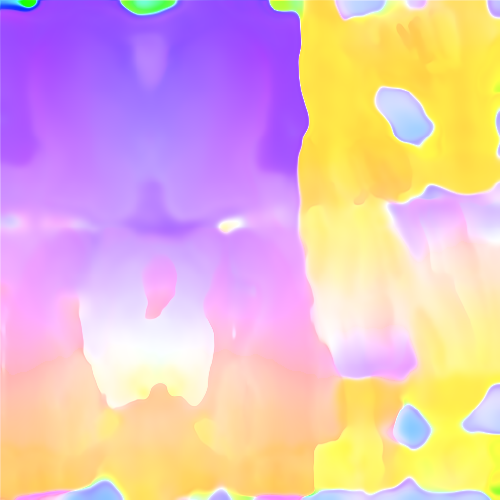}}}\\

\mpage{0.03}{\raisebox{0pt}{\rotatebox{90}{\small{UV texture/}}}}  \hfill
\hspace{-2mm}
\mpage{0.03}{\raisebox{0pt}{\rotatebox{90}{\small{Source image}}}}  \hfill
\hspace{-3mm}
\mpage{0.3}{\fbox{\includegraphics[width=\linewidth]{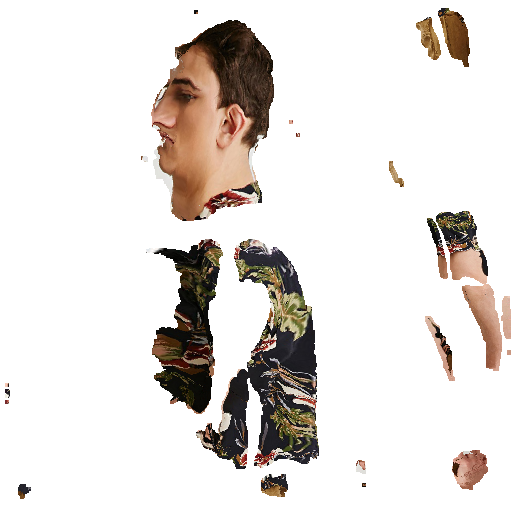}}\llap{\shiftleft{2.9cm}{\raisebox{1.6cm}{\includegraphics[height=1cm]{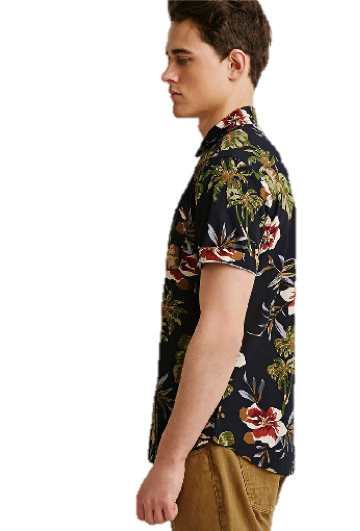}}}}}\hfill
\mpage{0.3}{\fbox{\includegraphics[width=\linewidth]{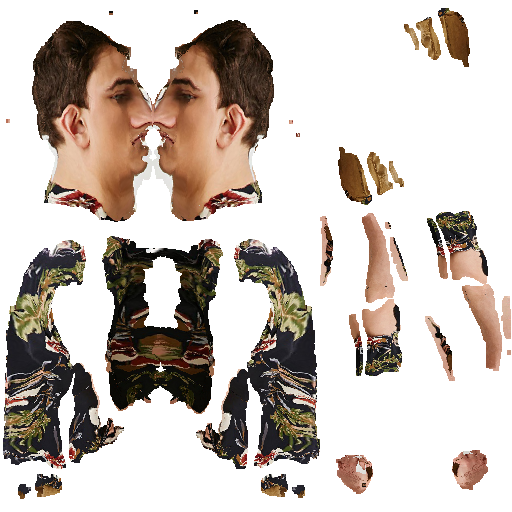}}}\hfill
\mpage{0.3}{\fbox{\includegraphics[width=\linewidth]{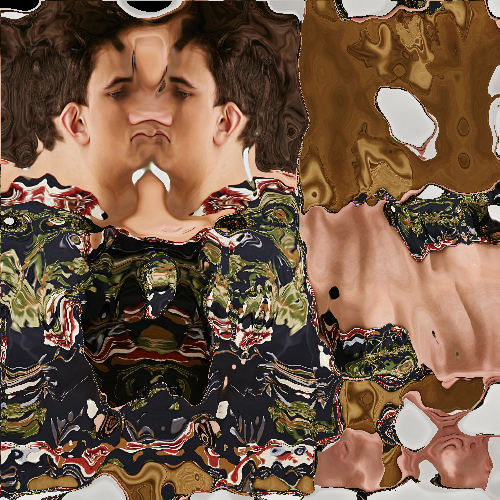}}}\\

}\hfill
\mpage{0.45}{

\mpage{0.48}{\includegraphics[width=\linewidth]{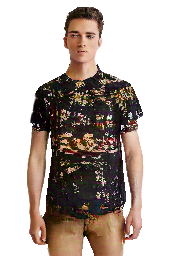}}\llap{\raisebox{-2.7cm}{\includegraphics[height=1.7cm]{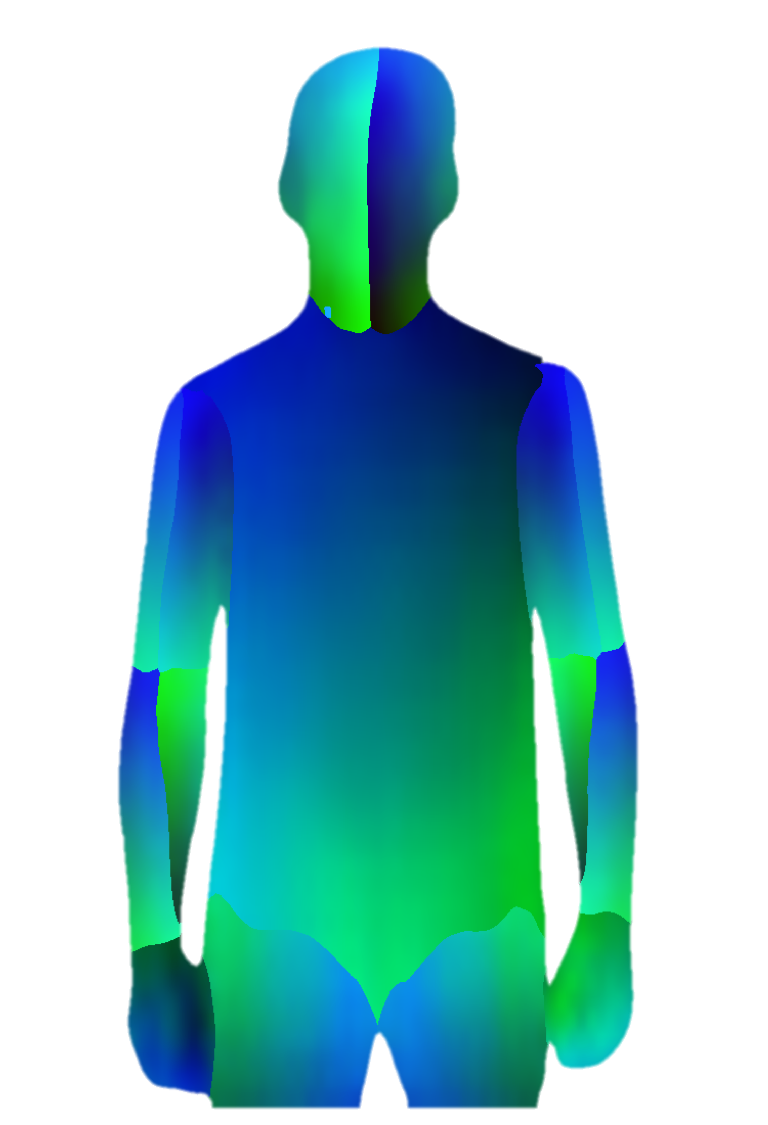}}}\hfill
\mpage{0.48}{\includegraphics[width=\linewidth]{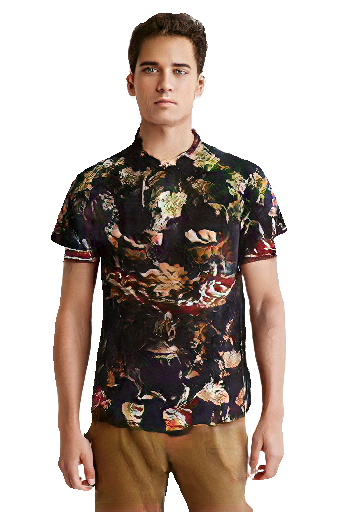}}\\
}\\

\mpage{0.52}{
\hspace{1mm}
\mpage{0.3}{\small{(a) Base}} \hfill
\mpage{0.3}{\small{(b) Base $\cup$ Mirrored}}\hfill
\mpage{0.33}{\small{(c) Sym.-guided inpainting}}\\}\hfill
\mpage{0.45}{
\vspace{1mm}
\mpage{0.48}{\small{(a) Base (without symmetry)}} \hfill
\mpage{0.48}{\small{(b) Ours (with symmetry)}}\\}\\

\mpage{0.53}{$\underbrace{\hspace{\textwidth}}_{\substack{\vspace{-7.0mm}\\\colorbox{white}{~~Symmetry-guided inpainting~~}}}$}\hfill
\mpage{0.45}{$\underbrace{\hspace{\textwidth}}_{\substack{\vspace{-7.0mm}\\\colorbox{white}{~~Reposed results~~}}}$}\\

\vspace{-3mm}
\caption{\textbf{Symmetry-guided inpainting.}
(\emph{Left}) Given a mesh grid and the source image dense pose $P_{src}$, we first map the coordinates from the 2D mesh grid to appropriate locations in the UV-space using a pre-computed mapping table.
We can then use these mapped base coordinates $C_{base}$ to warp RGB pixels from the input source image $I_{src}$.
We show the base coordinates and their warped RGB pixels in (a).
(b) In addition to these base coordinates, we can also map the left-right \emph{mirrored} coordinates $C_{mirrored}$ from the 2D mesh grid to the UV space. 
To train our coordinate completion model, we combine the incomplete base and mirrored coordinates in the UV-space.
We then concatenate these combined coordinates with their respective mask and pass them as input to our coordinate completion model.
We show our completed coordinates and the UV texture map in (c).
(\emph{Right}) We compare the reposing results \emph{without} and \emph{with} the proposed symmetry prior. 
}

\label{fig:symmetry}
\end{figure*}

\begin{figure}[t]
\centering
\mpage{0.48}{\includegraphics[width=\linewidth]{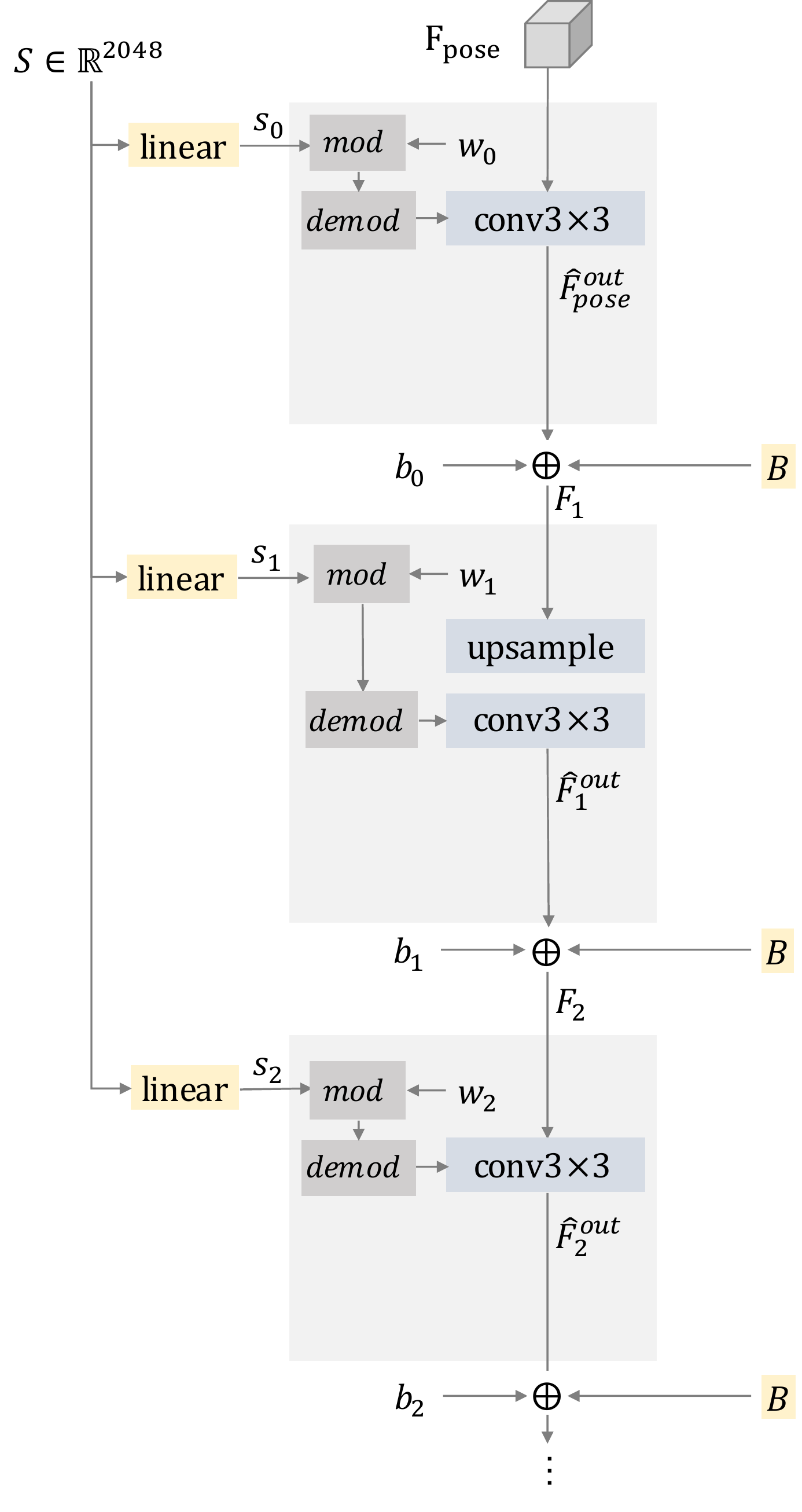}} \hfill
\mpage{0.48}{\includegraphics[width=\linewidth]{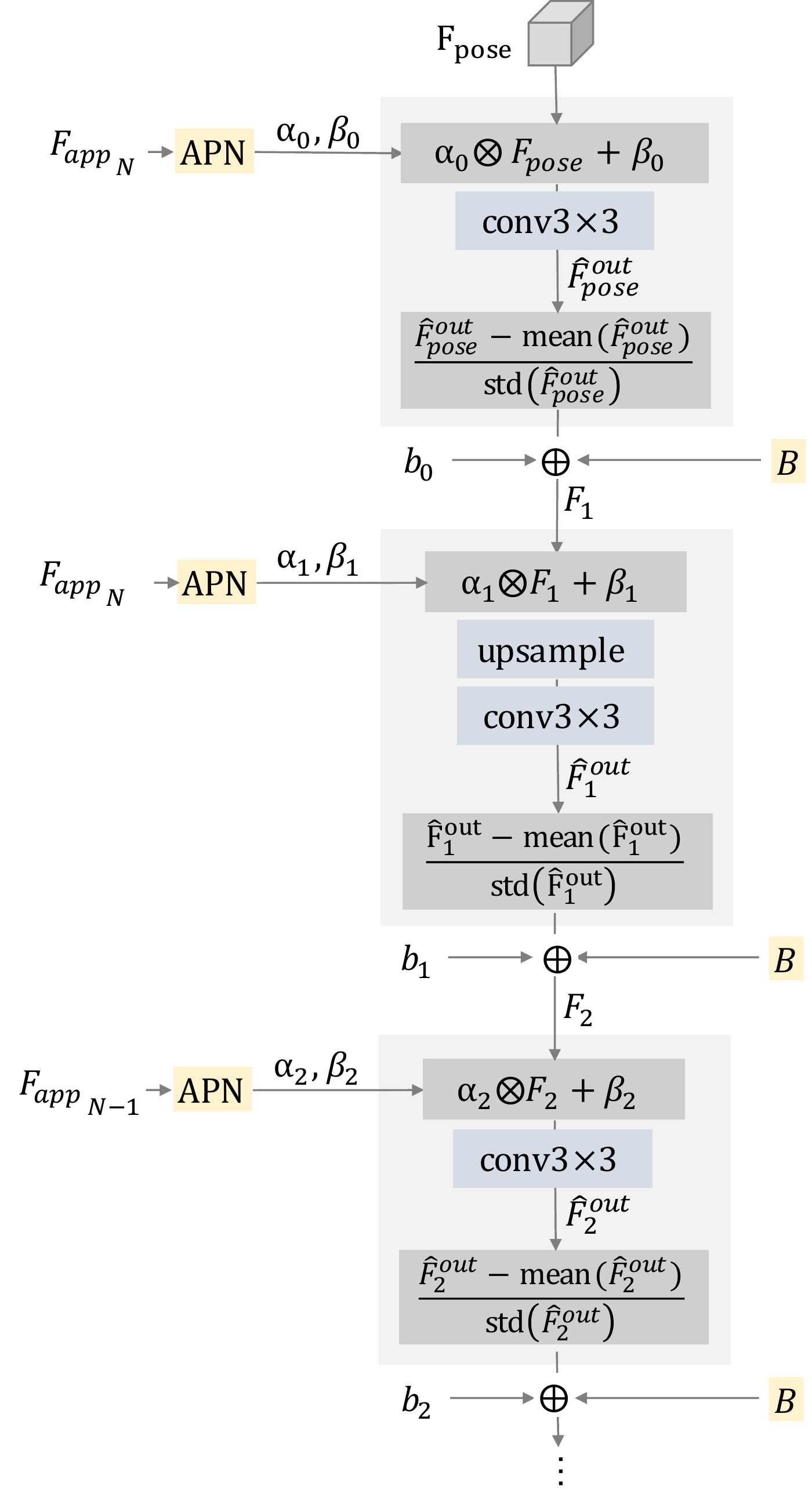}}
\\
\mpage{0.48}{\small{(a) Non-spatial modulation (StyleGAN2)}} \hfill
\mpage{0.48}{\small{(b) Spatial modulation \\(Ours)}}\\
\caption{\textbf{Spatial vs. non-spatial modulation of StyleGAN2 features.}
Our input to the StyleGAN2 generator is the encoded target pose features $F_{pose}$
(a) StyleGAN2~\cite{Karras2019stylegan2} performs \emph{non-spatial modulation} of features by modulating and demodulating the weights of the convolutions using the learned style vector $S$. 
After the convolution the bias is added as well as StyleGAN2 noise broadcast operation $B$.
(b) To better leverage the spatial features for preserving appearance details, we propose \emph{spatial modulation} of styleGAN2 features. 
Instead of modulating and demodulating the weights of the convolutions, we modulate the mean and standard deviation of the features. 
We perform this modulation before the convolution using the shifting and scaling parameters, $\alpha$ and $\beta$, generated by the affine parameters network (APN). 
We then normalize the output of the convolution to zero mean and unit standard deviation before adding the bias and StyleGAN2 noise broadcast operation $B$.
}
\label{fig:revised_modulation}
\end{figure}

\subsection{Source Feature Generator}
To preserve the appearance in source image $I_{src}$, we encode it using several residual blocks into multi-scale features $F_{app_i}^{src}$. 
We utilize the pretrained coordinate completion model to obtain the target image-space coordinates $T_{coord}$ such that it could warp the source features $F_{app_i}^{src}$ from the source pose to the target pose $F_{app_i}^{trg}$. 
We then concatenate these warped features with the target dense pose mask $M_{p_{trg}}$ and pass them into a feature pyramid network~\cite{lin2017feature} to get our multi-scale warped appearance features $F_{app_i}$. We show our source feature generator in \figref{sfg}.

\subsection{Affine Parameters Network and Spatial Modulation}
Prior to every convolution layer in each style block of StyleGAN2, we pass the warped source features $F_{app_i}$ into an affine parameters network to generate scaling $\alpha$, and shifting $\beta$ parameters. Each convolution layer has its own independent affine parameters network which is composed of two $1\times1$ convolutions separated by a ReLU activation function for each parameter. 
To preserve spatial details, we modify every convolution layer in each style block of StyleGAN2~\cite{Karras2019stylegan2}. 
Instead of performing spatially invariant weight modulation and demodulation, we use the generated scaling and shifting tensor parameters, $\alpha$ and $\beta$, to perform spatially varying modulation of the features $F_{i}$ as follows:

\begin{equation}\label{eqn:modulate}
\hat{F_\mathrm{i}} = \alpha_\mathrm{i} \otimes F_\mathrm{i} + \beta_\mathrm{i},
\end{equation}
where $\hat{F_\mathrm{i}}$ is now the modulated features that will be passed as input to the $3\times3$ convolution of styleGAN2 generator. 
The output features of the convolution $\hat{F_\mathrm{i}}^{out}$ is then normalized to zero mean and unit standard deviation. Such that:
\begin{equation}\label{eqn:normalize}
\bar{F_\mathrm{i}} = \frac{\hat{F_\mathrm{i}}^{out} - \mathrm{mean}(\hat{F_\mathrm{i}}^{out})}{\mathrm{std}(\hat{F_\mathrm{i}}^{out})}
\end{equation}

Similar to StyleGAN2, we then add the noise broadcast operation $B$ and the bias to $\bar{F_\mathrm{i}}$ to get $F_{i+1}$ which will be fed into the next convolution layer.
In Figure~\ref{fig:revised_modulation}, we illustrate our detailed spatial modulation approach as well as the non-spatial weight modulation and demodulation of StyleGAN2.

\subsection{Training Losses}
In addition to StyleGAN2 adversarial loss $L_{adv}$, we train our model to minimize the following reconstruction losses:

\topic{$\ell_1$ loss}
We minimize the $\ell_1$ loss between the foreground human regions of the synthesized image $\hat{I}_{trg}$ and of the ground truth target $I_{trg}$.
\begin{equation}\label{eqn:l1}
L_{\ell_1} = ||\hat{I}_{trg} \cdot M_{trg}  - I_{trg} \cdot M_{trg}||_{1},
\end{equation}
where $M_{trg}$ is the human foreground mask estimated using a human parsing method~\cite{gong2018instance}.

\topic{Perceptual loss}
We minimize the weighted sum of the $\ell_1$ loss between the pretrained VGG features of the synthesized $\hat{I_{trg}}$ foreground and the ground truth $I_{trg}$ foreground such that:
\begin{equation}\label{eqn:vgg}
L_{vgg} = \sum_{i=1}^{5} w_{i} \cdot ||VGG_{l_i}(\hat{I}_{trg}\cdot M_{trg})  - VGG_{l_i}(I_{trg} \cdot M_{trg})||_{1}
\end{equation}
We use $w = [\frac{1}{32}, \frac{1}{16}, \frac{1}{8}, \frac{1}{4}, 1.0]$ and VGG ReLU output layers $l=[1, 6, 11, 20, 29]$ following~\cite{wang2017high, park2019SPADE}.

\topic{Face identity loss}
We use MTCNN~\cite{zhang2016joint} to detect, crop, and align faces from the generated image $\hat{I}_{trg}$ and ground truth target $I_{trg}$. When a face is detected, we maximize the cosine similarity between the pretrained SphereFace~\cite{liu2017sphereface} features of the generated face and the ground truth target face. Such that:
\begin{equation}\label{eqn:face}
L_{face} = 1 - 
\left(\frac{SF(\hat{I}_{trg})^\top SF(I_{trg})}{\mathrm{max}(||SF(\hat{I}_{trg})||_2 \cdot ||SF(I_{trg})||_2, \epsilon)}\right)
\end{equation}
where SF is the pretrained SphereFace feature extractor, and $SF(\hat{I}_{trg})$ and $SF(I_{trg})$ are features of the aligned faces of the generated and ground truth image respectively. $\epsilon = e^{-8}$ is a very small value to avoid zero-dividing.

Therefore, our final loss is:
$L = L_{adv} + L_{\ell_1} + L_{vgg} + L_{face}$.

\section{Experimental Results}
\label{sec:results}

\subsection{Experimental setup}
\topic{Implementation details.}
We implement our model with PyTorch. 
We use ADAM optimizer with a learning rate of $ratio \cdot 0.002$ and beta parameters $(0, 0.99^{ratio})$. 
We set the generator ratio to $\frac{4}{5}$ and discriminator ratio to $\frac{16}{17}$. 

\topic{Training}
We first train our model by focusing on generating the foreground. We apply the reconstruction loss and the adversarial loss only on the foreground.%
We set the batch size to 1 and train for 50 epochs.
This training process takes around 7 days on 8 NVIDIA 2080 Ti GPUs. 
We then finetune the model by applying the adversarial loss globally on the entire image. We set the batch size to 8 and train for 10 epochs.
This training process takes less than 2 days on 2 A100 GPUs. 
At test time, generating a reposing results with $384 \times 512$ resolution takes 0.4 seconds using 1 NVIDIA 2080 Ti GPU. 

\topic{Dataset.}
We use the DeepFashion dataset~\cite{liuLQWTcvpr16DeepFashion} for training and evaluation. We follow the train/test splits (101,967 training and 8,570 testing pairs) of recent methods~\cite{ADGAN_2020, GFLA_2020, PATN_2019}.

\begin{table}[t]\setlength{\tabcolsep}{3pt}
	\centering%
\caption{
Quantitative comparison with the state-of-the-art methods on the DeepFashion dataset~\cite{liuLQWTcvpr16DeepFashion}.
\vspace{-3mm}
}
	\begin{tabular}{lcccc}
		\toprule
    	             	         & PSNR$\uparrow$    & SSIM$\uparrow$ & FID$\downarrow$ & LPIPS$\downarrow$\\
        \midrule
    	$Resolution = 174\times256$	     &                   &                &                   &        \\
        
		 PATN~\cite{PATN_2019}         & 17.7041           & 0.7543         & 21.8568         & 0.195     \\
         ADGAN~\cite{ADGAN_2020}       & 17.7223           & 0.7544         & 16.2686         & 0.175     \\
         GFLA~\cite{GFLA_2020}         & 18.0424           & 0.7625         & 15.1722         & 0.167     \\
         Ours&\textbf{18.5062}   &\textbf{0.7784} & \textbf{8.7449} & \textbf{0.134} \\ %
        \midrule
    	$Resolution = 348\times512$	     &                   &                &                   &        \\
        GFLA~\cite{GFLA_2020}         & 17.9718           & 0.7540         & 18.8519         & 0.170     \\
        Ours            &\textbf{18.3567}   &\textbf{0.7640}& \textbf{9.0002} & \textbf{0.143} \\ %
		\bottomrule
		\label{tab:sota}
	\end{tabular}
\end{table}

\begin{table}[t]\setlength{\tabcolsep}{1pt}
	\centering%
\caption{
Quantitative comparison on $348\times512$ resolution with StylePoseGAN~\cite{sarkar2021style} on their DeepFashion dataset train/test split.
}
\vspace{-3mm}
	\begin{tabular}{lcccc}
		\toprule
    	             	              & PSNR$\uparrow$& SSIM$\uparrow$ & FID$\downarrow$ & LPIPS$\downarrow$\\
        \midrule
        StylePoseGAN~\cite{sarkar2021style} & 17.7568     & 0.7508    & 7.4804   & 0.167     \\
        Ours  & \textbf{18.5029} & \textbf{0.7711} & \textbf{6.0557} & \textbf{0.144}     \\ %
		\bottomrule
		\label{tab:sarkar}
	\end{tabular}
\end{table}

\begin{figure}[t]
\centering

\mpage{0.17}{\includegraphics[width=\linewidth, trim=27 0 27 0, clip]{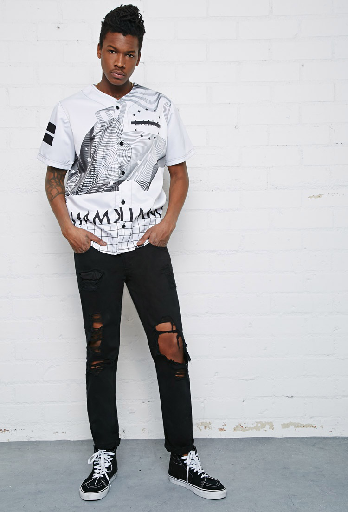}\llap{\includegraphics[height=1.1cm]{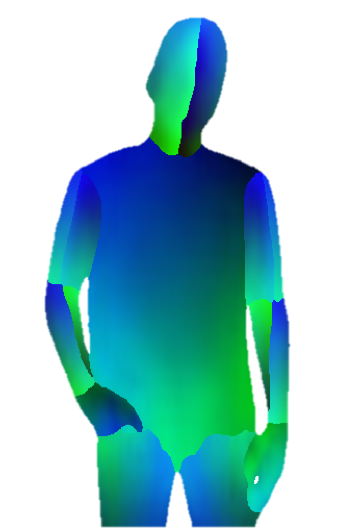}}}\hfill
\hspace{-4mm}
\mpage{0.17}{\includegraphics[width=\linewidth, trim=15 0 15 0, clip]{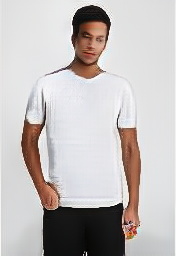}}\hfill
\hspace{-4mm}
\mpage{0.17}{\includegraphics[width=\linewidth, trim=15 0 15 0, clip]{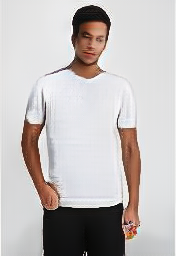}}\hfill
\hspace{-4mm}
\mpage{0.17}{\includegraphics[width=\linewidth, trim=30 0 30 0, clip]{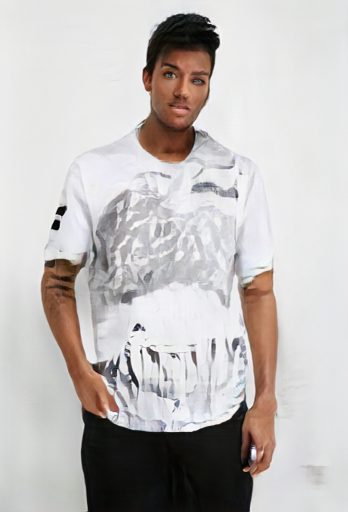}}\hfill
\hspace{-4mm}
\mpage{0.17}{\includegraphics[width=\linewidth, trim=30 0 30 0, clip]{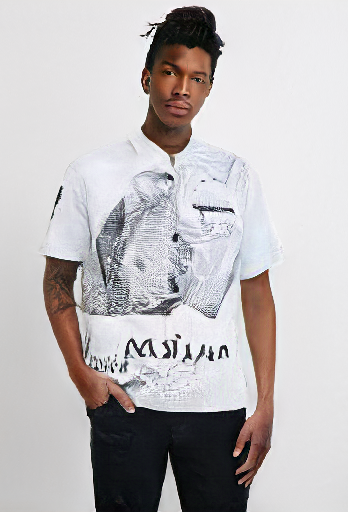}}\hfill
\hspace{-4mm}
\mpage{0.17}{\includegraphics[width=\linewidth, trim=30 0 30 0, clip]{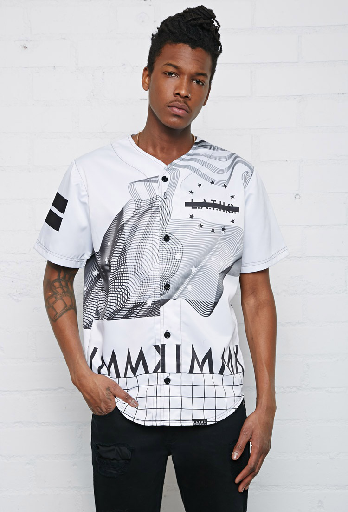}}\\
\mpage{0.17}{\includegraphics[width=\linewidth, trim=27 0 27 0, clip]{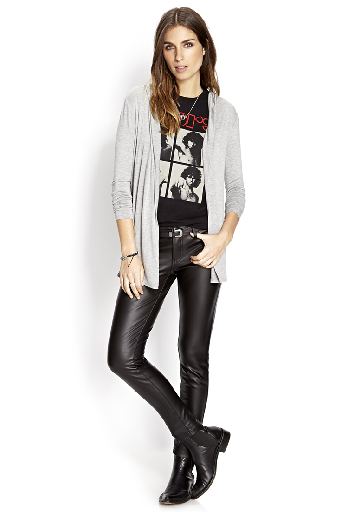}\llap{\includegraphics[height=1.1cm]{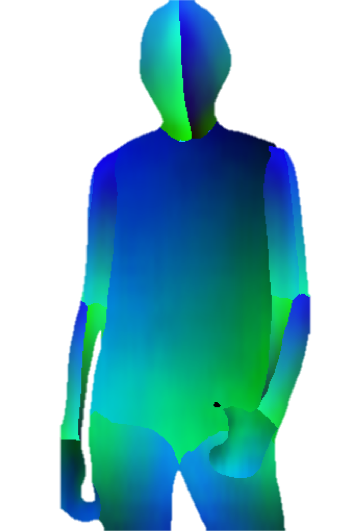}}}\hfill
\hspace{-4mm}
\mpage{0.17}{\includegraphics[width=\linewidth, trim=15 0 15 0, clip]{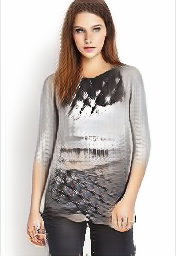}}\hfill
\hspace{-4mm}
\mpage{0.17}{\includegraphics[width=\linewidth, trim=15 0 15 0, clip]{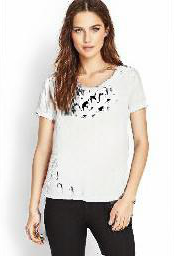}}\hfill
\hspace{-4mm}
\mpage{0.17}{\includegraphics[width=\linewidth, trim=30 0 30 0, clip]{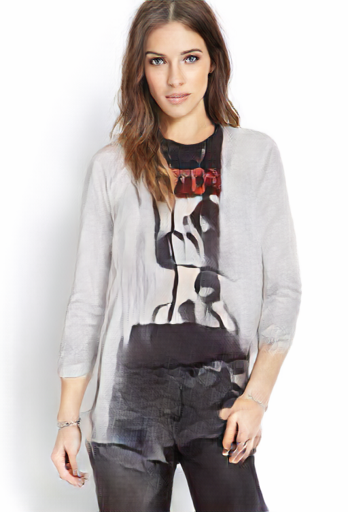}}\hfill
\hspace{-4mm}
\mpage{0.17}{\includegraphics[width=\linewidth, trim=30 0 30 0, clip]{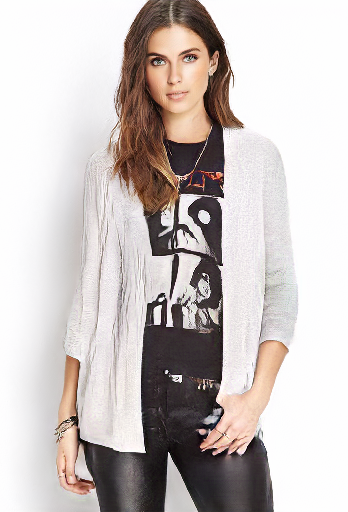}}\hfill
\hspace{-4mm}
\mpage{0.17}{\includegraphics[width=\linewidth, trim=30 0 30 0, clip]{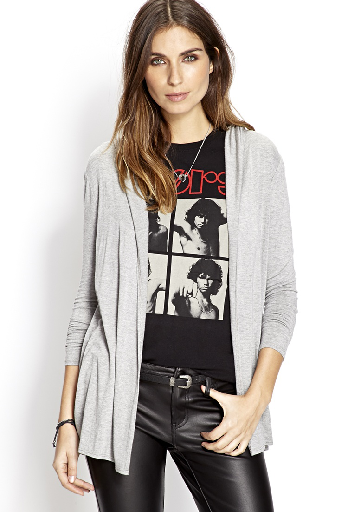}}\\
\mpage{0.17}{\includegraphics[width=\linewidth, trim=27 0 27 0, clip]{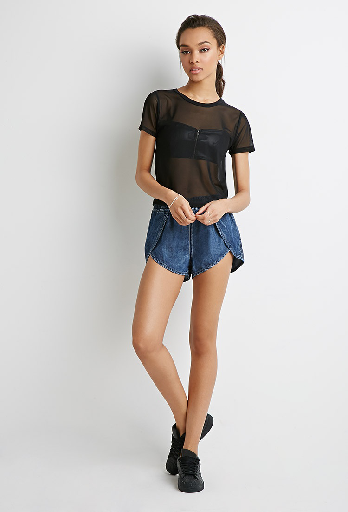}\llap{\includegraphics[height=1.1cm]{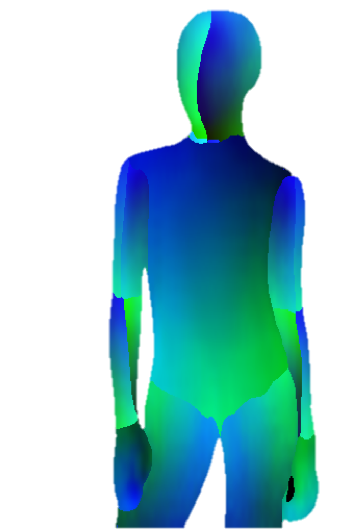}}}\hfill
\hspace{-4mm}
\mpage{0.17}{\includegraphics[width=\linewidth, trim=15 0 15 0, clip]{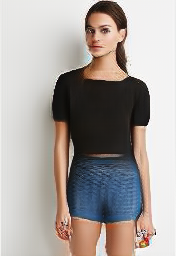}}\hfill
\hspace{-4mm}
\mpage{0.17}{\includegraphics[width=\linewidth, trim=15 0 15 0, clip]{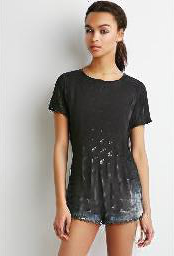}}\hfill
\hspace{-4mm}
\mpage{0.17}{\includegraphics[width=\linewidth, trim=30 0 30 0, clip]{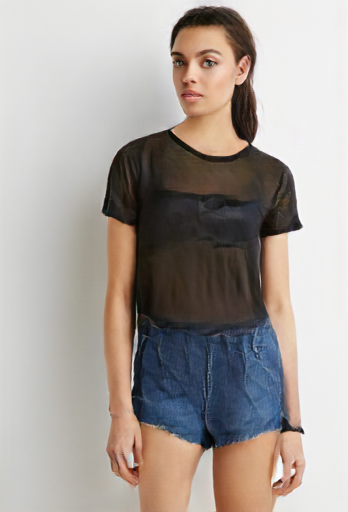}}\hfill
\hspace{-4mm}
\mpage{0.17}{\includegraphics[width=\linewidth, trim=30 0 30 0, clip]{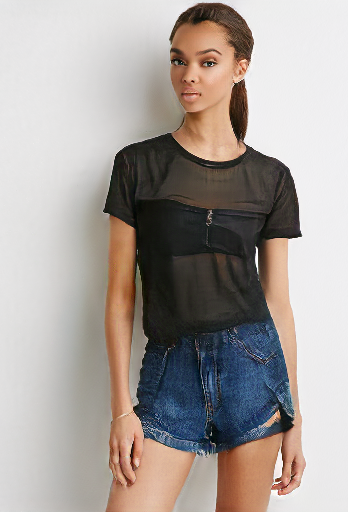}}\hfill
\hspace{-4mm}
\mpage{0.17}{\includegraphics[width=\linewidth, trim=30 0 30 0, clip]{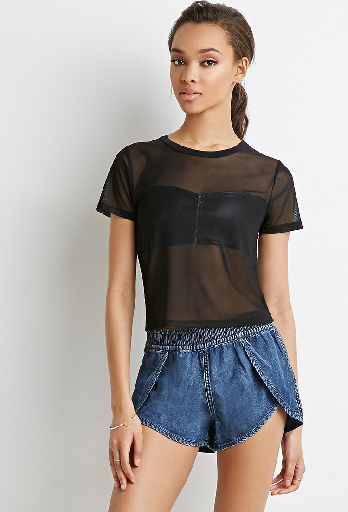}}\\
\mpage{0.17}{\includegraphics[width=\linewidth, trim=27 0 27 0, clip]{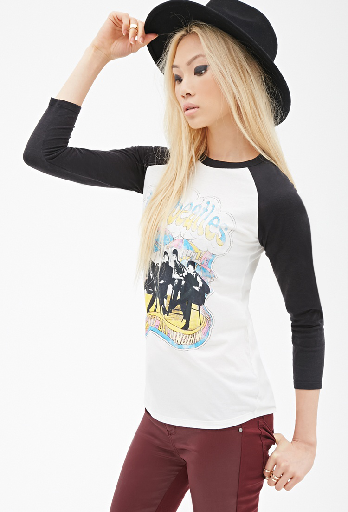}\llap{\includegraphics[height=1.1cm]{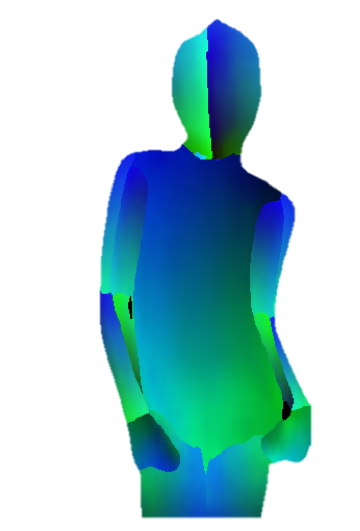}}}\hfill
\hspace{-4mm}
\mpage{0.17}{\includegraphics[width=\linewidth, trim=15 0 15 0, clip]{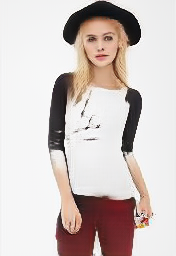}}\hfill
\hspace{-4mm}
\mpage{0.17}{\includegraphics[width=\linewidth, trim=15 0 15 0, clip]{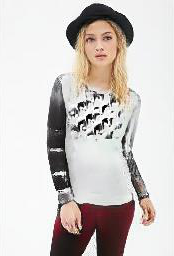}}\hfill
\hspace{-4mm}
\mpage{0.17}{\includegraphics[width=\linewidth, trim=30 0 30 0, clip]{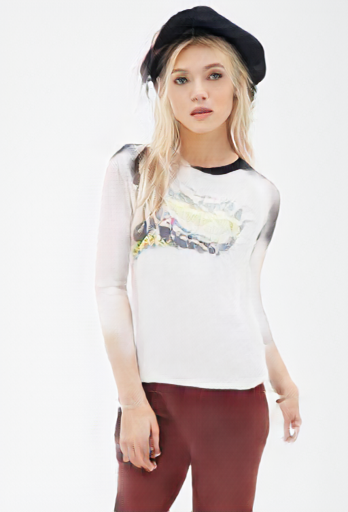}}\hfill
\hspace{-4mm}
\mpage{0.17}{\includegraphics[width=\linewidth, trim=30 0 30 0, clip]{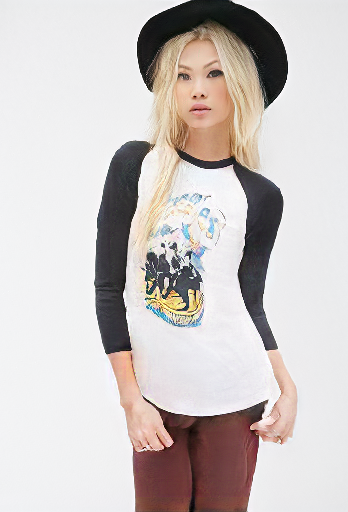}}\hfill
\hspace{-4mm}
\mpage{0.17}{\includegraphics[width=\linewidth, trim=30 0 30 0, clip]{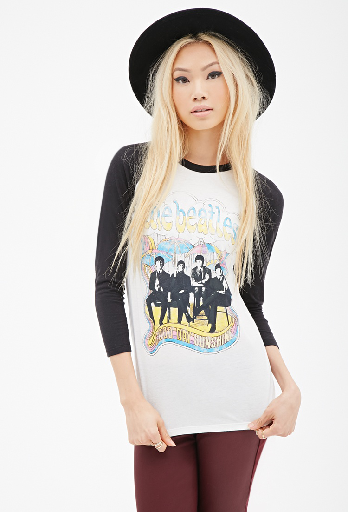}}\\
\mpage{0.17}{\small{Source/Pose}}\hfill
\hspace{-4mm}
\mpage{0.17}{\small{PATN}}\hfill
\hspace{-4mm}
\mpage{0.17}{\small{ADGAN}}\hfill
\hspace{-4mm}
\mpage{0.17}{\small{GFLA}}\hfill
\hspace{-4mm}
\mpage{0.17}{\small{Ours}}\hfill
\hspace{-4mm}
\mpage{0.17}{\small{Target}}\\

\vspace{-2mm}
\captionof{figure}{
\textbf{Visual comparison for human reposing.}
We show visual comparison of our 
method with  PATN~\cite{PATN_2019}, ADGAN~\cite{ADGAN_2020}, and GFLA~\cite{GFLA_2020} on DeepFashion dataset~\cite{liuLQWTcvpr16DeepFashion}.
Our approach successfully captures the local details from the source image.
}

\label{fig:sota}
\end{figure}
\begin{figure}[t]
\centering

\mpage{0.24}{\includegraphics[width=\linewidth, trim=27 0 27 0, clip]{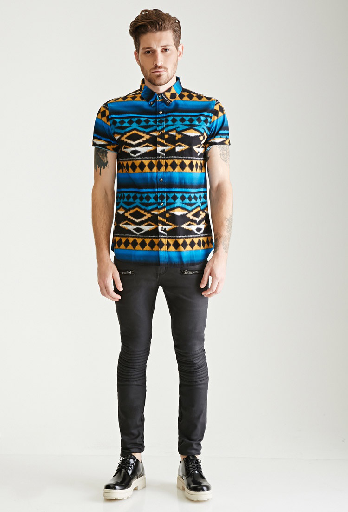}\llap{\includegraphics[height=1.3cm]{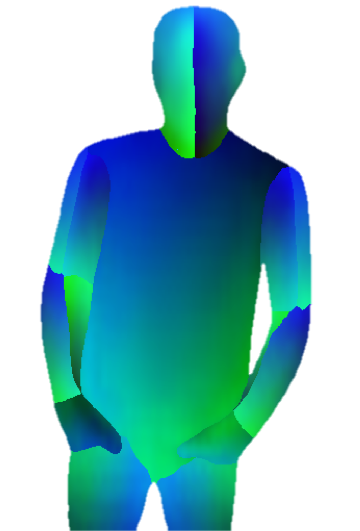}}}\hfill
\hspace{-4mm}
\mpage{0.24}{\includegraphics[width=\linewidth, trim=30 0 30 0, clip]{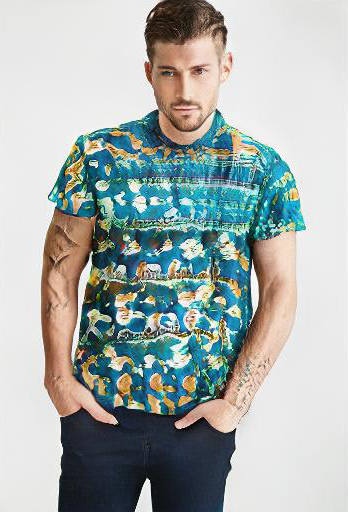}}\hfill
\hspace{-4mm}
\mpage{0.24}{\includegraphics[width=\linewidth, trim=30 0 30 0, clip]{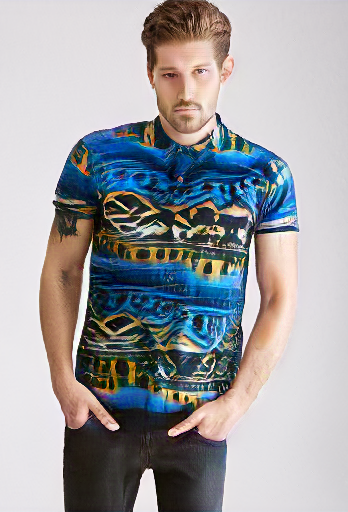}}\hfill
\hspace{-4mm}
\mpage{0.24}{\includegraphics[width=\linewidth, trim=30 0 30 0, clip]{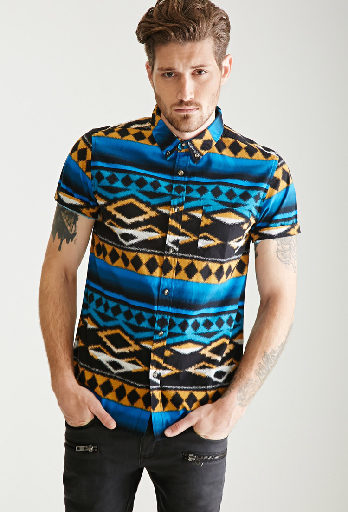}}\\
\vspace{-1mm}
\mpage{0.24}{\includegraphics[width=\linewidth, trim=27 0 27 0, clip]{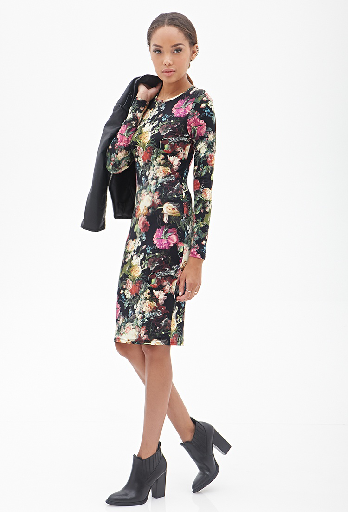}\llap{\includegraphics[height=1.3cm]{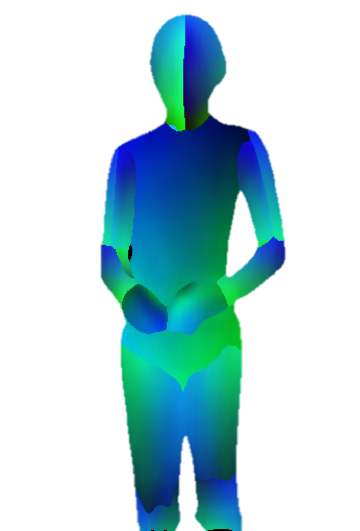}}}\hfill
\hspace{-4mm}
\mpage{0.24}{\includegraphics[width=\linewidth, trim=30 0 30 0, clip]{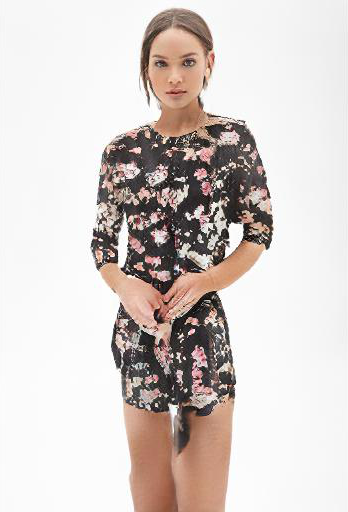}}\hfill
\hspace{-4mm}
\mpage{0.24}{\includegraphics[width=\linewidth, trim=30 0 30 0, clip]{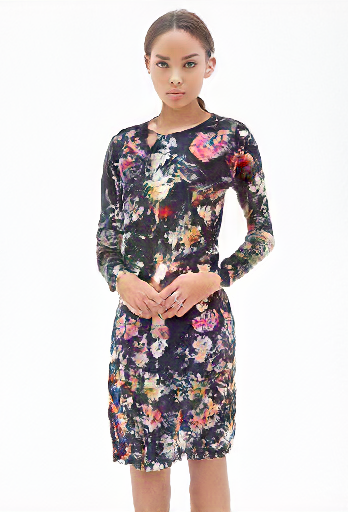}}\hfill
\hspace{-4mm}
\mpage{0.24}{\includegraphics[width=\linewidth, trim=30 0 30 0, clip]{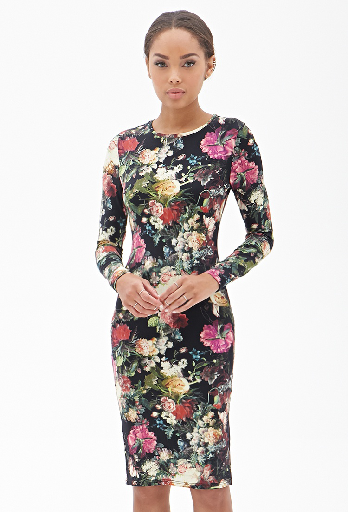}}\\
\vspace{-1mm}
\mpage{0.24}{\includegraphics[width=\linewidth, trim=27 0 27 0, clip]{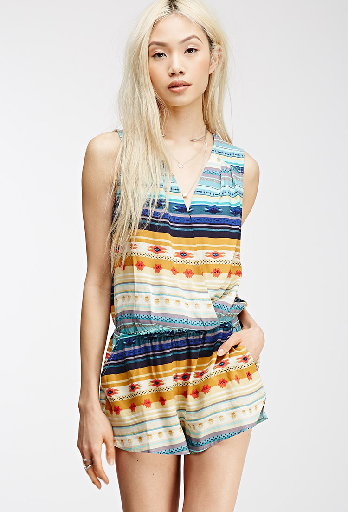}\llap{\includegraphics[height=1.3cm]{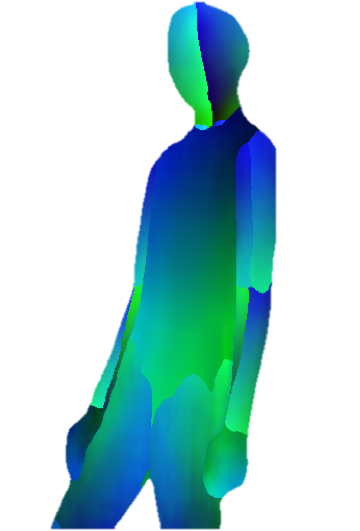}}}\hfill
\hspace{-4mm}
\mpage{0.24}{\includegraphics[width=\linewidth, trim=30 0 30 0, clip]{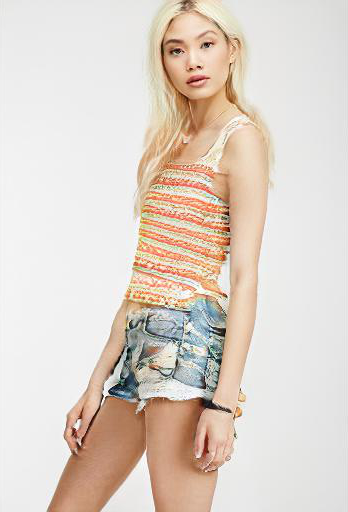}}\hfill
\hspace{-4mm}
\mpage{0.24}{\includegraphics[width=\linewidth, trim=30 0 30 0, clip]{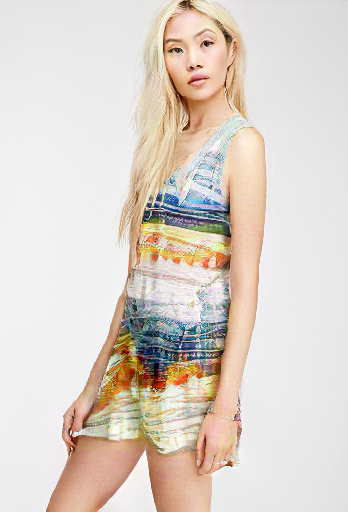}}\hfill
\hspace{-4mm}
\mpage{0.24}{\includegraphics[width=\linewidth, trim=30 0 30 0, clip]{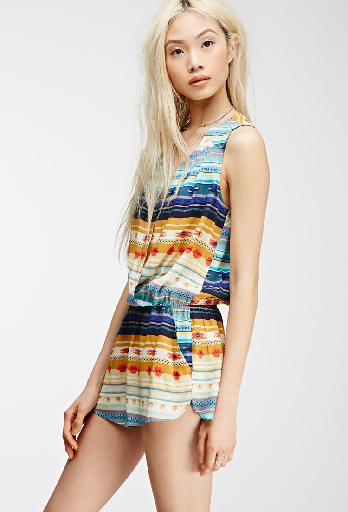}}\\
\mpage{0.24}{\small{Source/Pose}}\hfill
\hspace{-4mm}
\mpage{0.24}{\small{StylePoseGAN}}\hfill
\hspace{-4mm}
\mpage{0.24}{\small{Ours}}\hfill
\hspace{-4mm}
\mpage{0.24}{\small{Target}}\\

\vspace{-2mm}
\captionof{figure}{
\textbf{Visual comparison for human reposing.}
We compare our method with StylePoseGAN~\cite{sarkar2021style} on their train/test split of DeepFashion dataset~\cite{liuLQWTcvpr16DeepFashion}.
Our approach preserves the appearance and captures the fine-grained details of the source image.
}

\label{fig:sarkar}
\end{figure}

\subsection{Evaluations}
\topic{Quantitative evaluation} is reported in Table~\ref{tab:sota}. We report the human foreground peak signal-to-noise ratio (PSNR), structural similarity index measure (SSIM), learned perceptual image patch similarity (LPIPS)~\cite{lpips}, and Frechet Inception Distance (FID)~\cite{fid}. 
PSNR/SSIM often do not correlate well with perceived quality, particularly for synthesis tasks. For example, PSNR may favor blurry results over sharp ones. We report these metrics only for completeness.

Our method compares favorably against existing works such as PATN~\cite{PATN_2019} ADGAN~\cite{ADGAN_2020}, and GFLA~\cite{GFLA_2020}. 

Our method also compares favorably against the concurrent work StylePoseGAN~\cite{sarkar2021style}. We report the quantitative evaluation in Table~\ref{tab:sarkar}. We train and test our method using their DeepFashion dataset train/test split.%

\topic{Visual comparison} in Figure~\ref{fig:sota} and Figure~\ref{fig:sarkar} show that our proposed approach captures finer-grained appearance details from the input source images.

\begin{table}[t]\setlength{\tabcolsep}{12pt}
	\centering%
\caption{
The ability to preserve the identity of the reposed person.
}
\vspace{-3mm}
	\begin{tabular}{lcc}
		\toprule
    	    & Arcface & Spherenet \\
        \midrule

        GFLA~\cite{GFLA_2020}   & 0.117           & 0.197       \\
        Ours                    & \underline{0.373}          & \underline{0.438}       \\
        Ground truth            & \textbf{0.555}           & \textbf{0.444}        \\
		\bottomrule
		\label{tab:face_id}
	\end{tabular}
\end{table}

\topic{Face identity} 
We evaluate our model's ability to preserve the reposed person's identity. 
For test images with visible faces (7,164 from 8,570), we report in Table~\ref{tab:face_id} the averaged cosine similarity between the face features (Arcface~\cite{deng2019arcface} and Spherenet~\cite{coors2018spherenet}) extracted from the aligned faces in the source/target images. Our method compares favorably against GFLA~\cite{GFLA_2020}.%

\begin{figure*}[t]
\centering

\mpage{0.115}{\includegraphics[width=\linewidth, trim=50 0 50 0, clip]{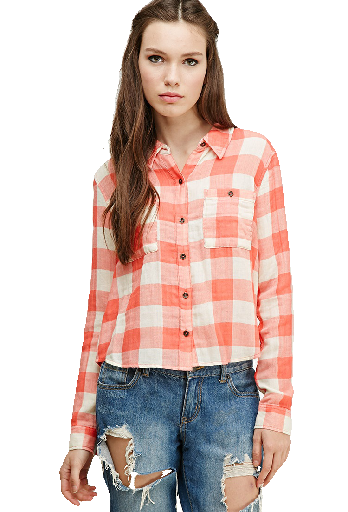}\llap{\includegraphics[height=1.75cm]{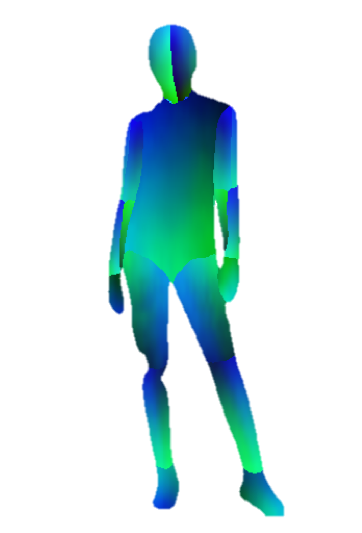}}}\hfill
\mpage{0.115}{\includegraphics[width=\linewidth, trim=60 0 60 0, clip]{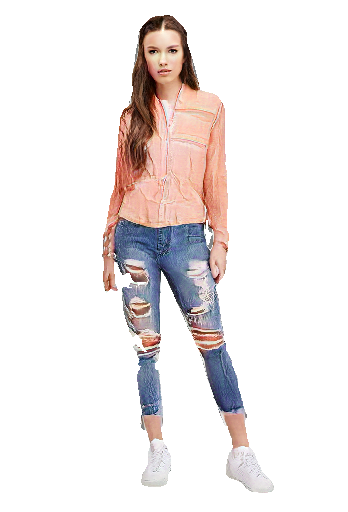}}\hfill
\mpage{0.115}{\includegraphics[width=\linewidth, trim=60 0 60 0, clip]{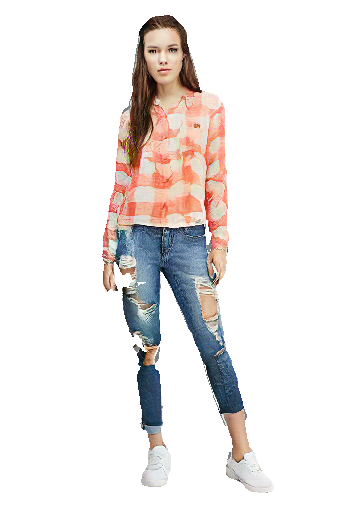}}\hfill
\mpage{0.115}{\includegraphics[width=\linewidth, trim=60 0 60 0, clip]{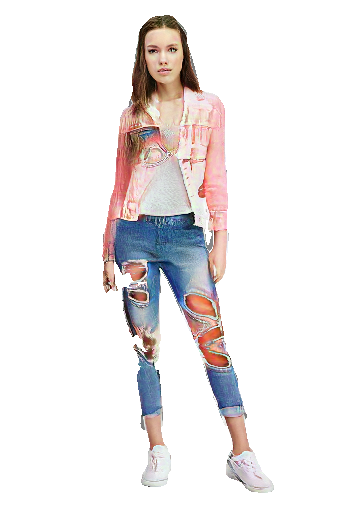}}\hfill
\mpage{0.115}{\includegraphics[width=\linewidth, trim=60 0 60 0, clip]{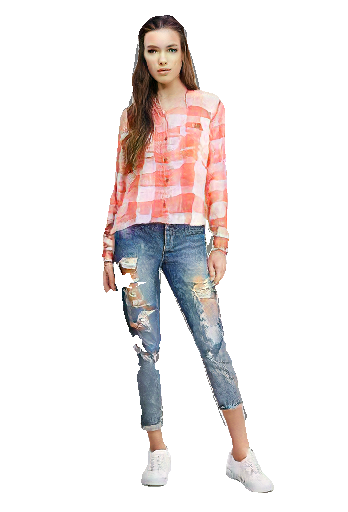}}\hfill
\mpage{0.115}{\includegraphics[width=\linewidth, trim=60 0 60 0, clip]{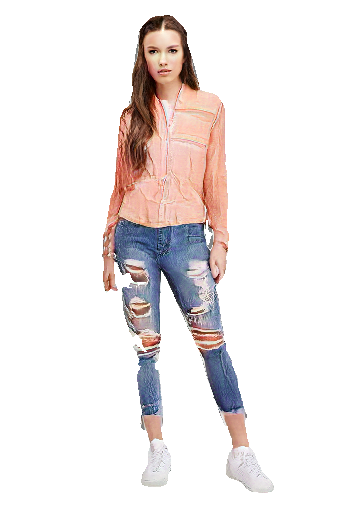}}\hfill
\mpage{0.115}{\includegraphics[width=\linewidth, trim=60 0 60 0, clip]{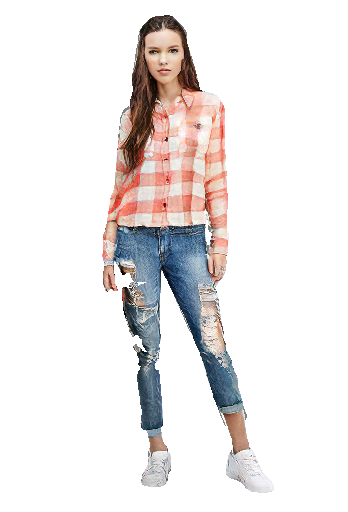}}\hfill
\mpage{0.115}{\includegraphics[width=\linewidth, trim=60 0 60 0, clip]{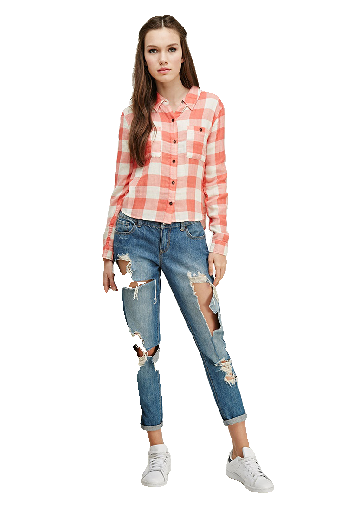}}\hfill
\\
\vspace{-4mm}
\mpage{0.115}{\includegraphics[width=\linewidth, trim=50 0 50 0, clip]{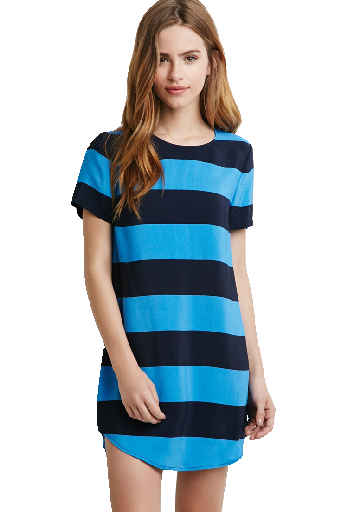}\llap{\includegraphics[height=1.75cm]{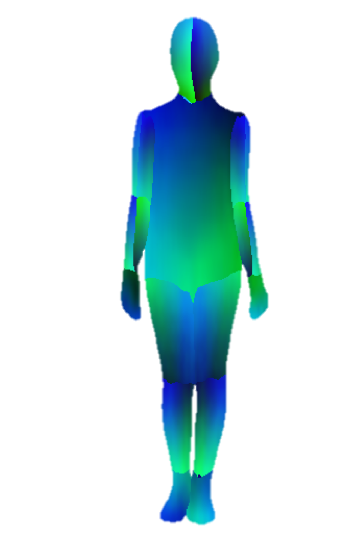}}}\hfill
\mpage{0.115}{\includegraphics[width=\linewidth, trim=60 0 60 0, clip]{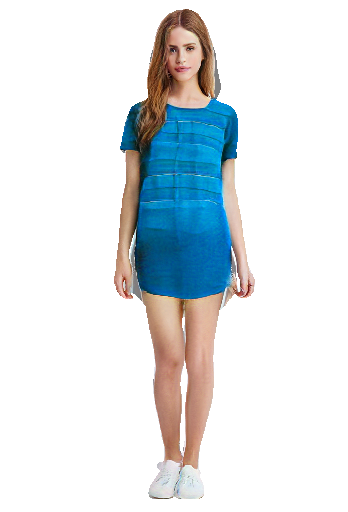}}\hfill
\mpage{0.115}{\includegraphics[width=\linewidth, trim=60 0 60 0, clip]{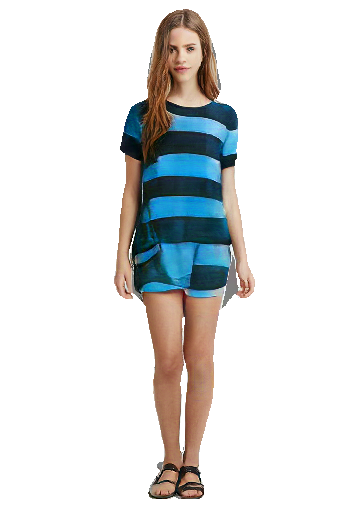}}\hfill
\mpage{0.115}{\includegraphics[width=\linewidth, trim=60 0 60 0, clip]{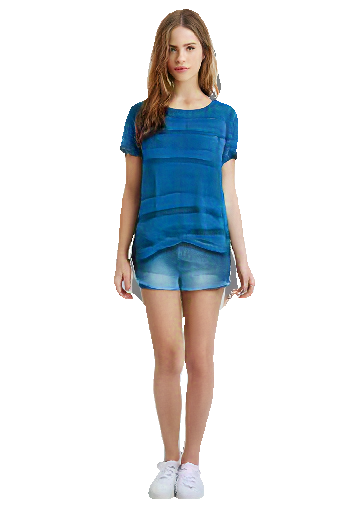}}\hfill
\mpage{0.115}{\includegraphics[width=\linewidth, trim=60 0 60 0, clip]{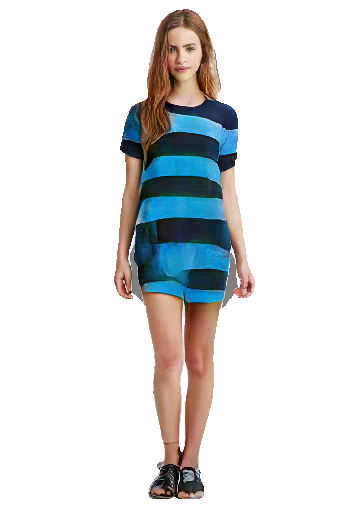}}\hfill
\mpage{0.115}{\includegraphics[width=\linewidth, trim=60 0 60 0, clip]{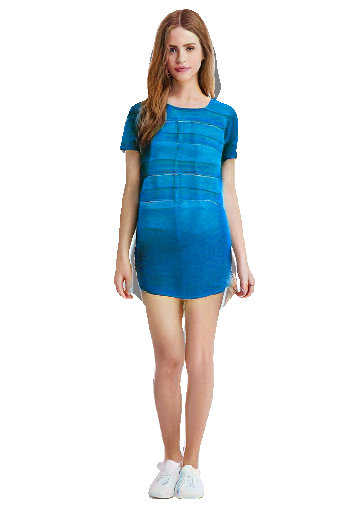}}\hfill
\mpage{0.115}{\includegraphics[width=\linewidth, trim=60 0 60 0, clip]{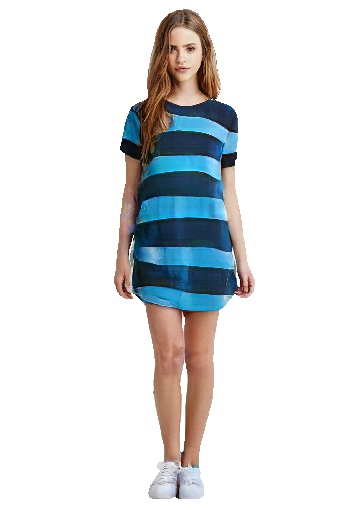}}\hfill
\mpage{0.115}{\includegraphics[width=\linewidth, trim=60 0 60 0, clip]{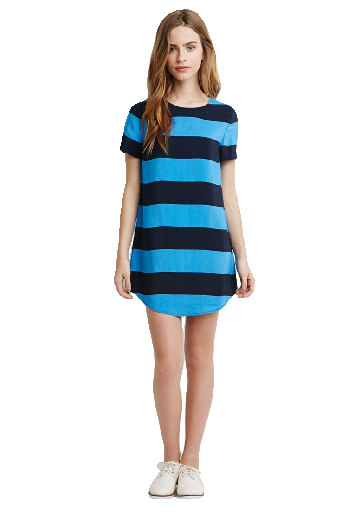}}\hfill
\\

\mpage{0.115}{Source/Pose}\hfill
\mpage{0.115}{A \\ non-spatial }\hfill  %
\mpage{0.115}{B \\ spatial}\hfill  %
\mpage{0.115}{C \\ non-spatial}\hfill  %
\mpage{0.115}{D \\ spatial}\hfill  %
\mpage{0.115}{E \\ non-spatial}\hfill  %
\mpage{0.115}{F \\ spatial}\hfill    %
\mpage{0.115}{Target}\\

\mpage{0.115}{\hspace{\textwidth}}\hfill
\mpage{0.23}{$\underbrace{\hspace{\textwidth}}_{\substack{\vspace{-7.0mm}\\\colorbox{white}{~~Incomplete UV~~}}}$}\hfill
\mpage{0.23}{$\underbrace{\hspace{\textwidth}}_{\substack{\vspace{-7.0mm}\\\colorbox{white}{~~Completed UV~~}}}$}\hfill
\mpage{0.23}{$\underbrace{\hspace{\textwidth}}_{\substack{\vspace{-7.0mm}\\\colorbox{white}{~~Source image foreground~~}}}$}\hfill
\mpage{0.115}{\hspace{\textwidth}}\\

\vspace{-5mm}
\caption{\textbf{Ablation.}
We compare our results with other variants, including the modulation types and source of appearance features.
We show that the proposed spatial modulation captures finer-grained details from the source image.
Transferring appearance features from the source image leads to fewer artifacts compared to features from the UV space. 
}
\vspace{-1mm}
\label{fig:ablation}
\end{figure*}

\subsection{Ablation study}
For the ablation study, we report the results of the foreground-focused trained model.

\topic{Symmetry prior.}
We evaluate the effectiveness of adding the symmetry prior to the input of the coordinate completion model. 
We train our networks \emph{with} and \emph{without} this symmetry prior and report the quantitative results in Table~\ref{tab:symmetry}. 
Results show that adding the symmetry prior indeed improves the quality of the synthesis. We note that the symmetry prior generally works well for repetitive/textured patterns, but may introduce artifacts for unique patterns (e.g., text).
\begin{table}[t]\setlength{\tabcolsep}{3.5pt}
	\centering%
\caption{
The effect of symmetry-guided coordindate inpainting on the DeepFashion dataset \cite{liuLQWTcvpr16DeepFashion}.
}
\vspace{-3mm}
	\begin{tabular}{lcccc}
		\toprule
    	             	         & PSNR$\uparrow$    & SSIM$\uparrow$ & FID$\downarrow$ & LPIPS$\downarrow$\\
        \midrule
        Without symmetry         & 18.8810           & 0.7886         & 8.5240         & 0.129          \\
        With symmetry (Ours)     &\textbf{18.9657}   &\textbf{0.7919} & \textbf{8.1434} & \textbf{0.124} \\
		\bottomrule
		\label{tab:symmetry}
	\end{tabular}
	\vspace{-3mm}
\end{table}

\topic{Modulation schemes.}
We quantitatively and qualitatively demonstrate the effectiveness of our proposed spatial modulation. 
We show quantitative evaluation in Table~\ref{tab:ablation}. 
We show qualitative results in Figure~\ref{fig:ablation}. 
Results show that spatially varying modulation improves the quality of the synthesized human foreground and captures the spatial details of the source images regardless of the input source image type.
\begin{table}[t]\setlength{\tabcolsep}{3pt}
	\centering
\caption{Ablation on source for appearance (Incomplete UV, Complete UV, and Image) and modulation types (Spatial and Non-spatial).
}
\vspace{-3mm}
	\begin{tabular}{llccccc}
		\toprule
    	ID&Input Source  & Spatial	    & PSNR$\uparrow$    & SSIM$\uparrow$ & FID$\downarrow$ & LPIPS$\downarrow$\\       
        \midrule
		A&Incomplete UV  &     -      & 18.7385           & 0.7710         & 9.4150          & 0.151          \\
		B&Incomplete UV  & \checkmark  & 18.6005           & 0.7696         & 9.2321          & 0.146          \\
        C&Complete UV    &     -       &\textbf{18.9407}   &\textbf{0.7770} & 9.7435          & 0.147          \\
        D&Complete UV    & \checkmark  & 18.7063           & 0.7720        &\underline{9.0236}&\underline{0.143}\\
        E&Source image       &     -      & 18.6027           & 0.7678         & 9.4367         & 0.154           \\
        F&Source image & \checkmark  &\underline{18.7420}&\underline{0.7739} & \textbf{8.8060} & \textbf{0.139} \\
		\bottomrule
		\label{tab:ablation}
	\end{tabular}
	\vspace{-6mm}
\end{table}

\topic{Sources of appearance.}
We experiment with multiple variants for encoding source appearance. 
Specifically, the \emph{incomplete UV},  \emph{complete UV} (completed using the coordinate completion model), and \emph{source image} (our approach shown in \figref{revised_overview}).
We report the quantitative results in Table~\ref{tab:ablation} and visual results in Figure~\ref{fig:ablation}.
The results show that 
extracting appearance features directly from the source image preserves more details than other variants.

\subsection{Garment transfer Results}
\begin{figure}[t]
\centering
\includegraphics[width=0.95\linewidth]{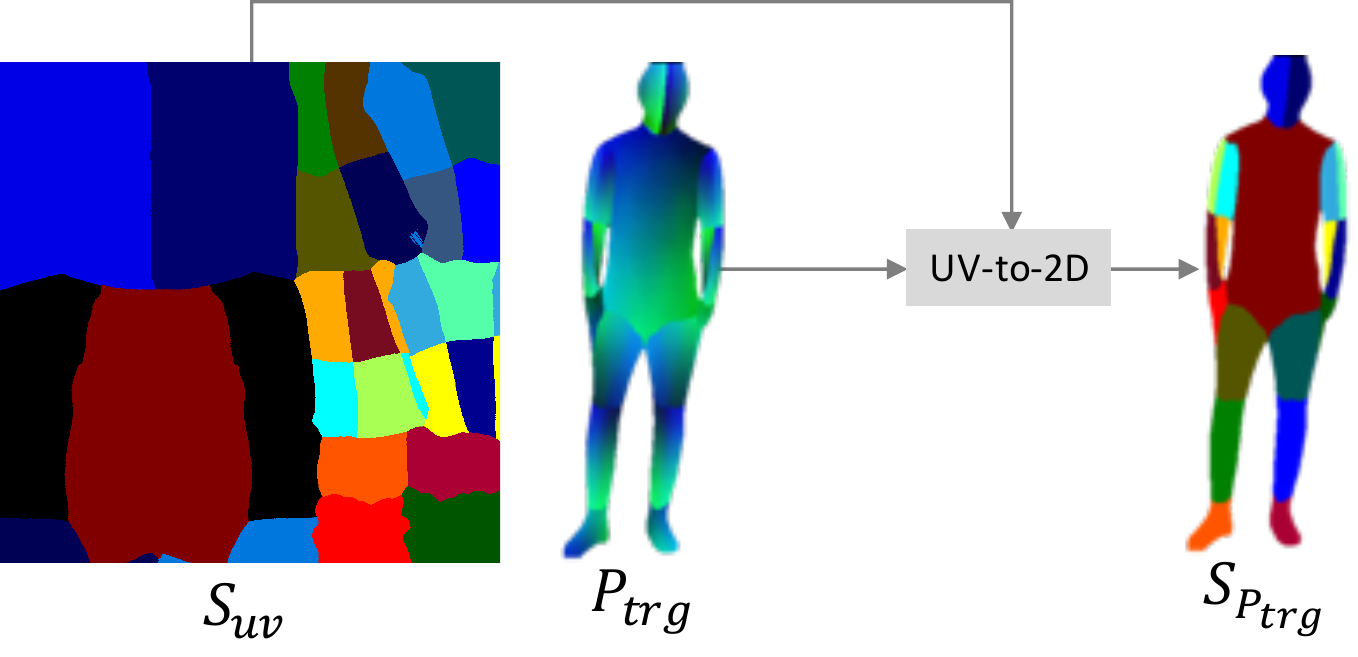}
\vspace{-4mm}
\caption{\textbf{Human body segmentation.}
We create a UV-space segmentation map $S_{uv}$ of the human body using the UV-space pre-computed mapping table. 
We use the target dense pose $P_t$ to map this UV-space segmentation map to 2D target pose $S_{P_t}$ which can then be used to combine features from multiple source images to perform garment transfer.
}
\vspace{-2mm}
\label{fig:uv_parts}
\end{figure}

\begin{figure}[t]
\centering
\mpage{0.03}{\raisebox{0pt}{\rotatebox{90}{\small{Source image/Target garment}}}}  \hfill
\hspace{-3mm}
\mpage{0.23}{\includegraphics[width=\linewidth, trim=65 0 65 0, clip]{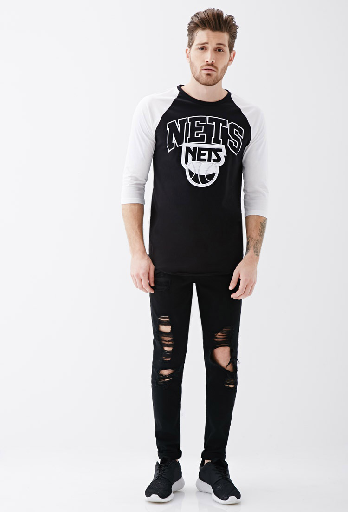}\llap{\includegraphics[height=1.75cm]{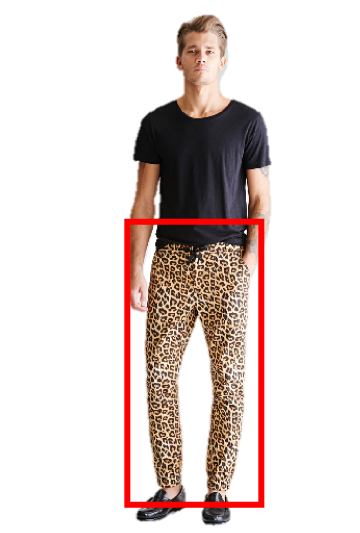}}}\hfill
\mpage{0.23}{\includegraphics[width=\linewidth, trim=65 0 65 0, clip]{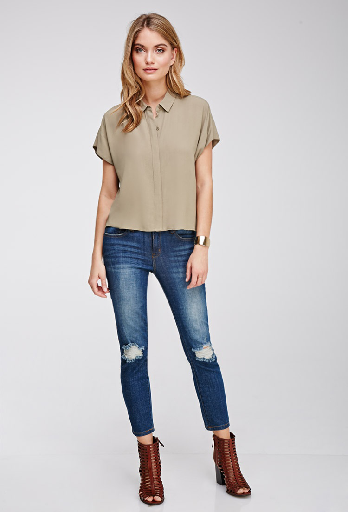}\llap{\includegraphics[height=1.75cm]{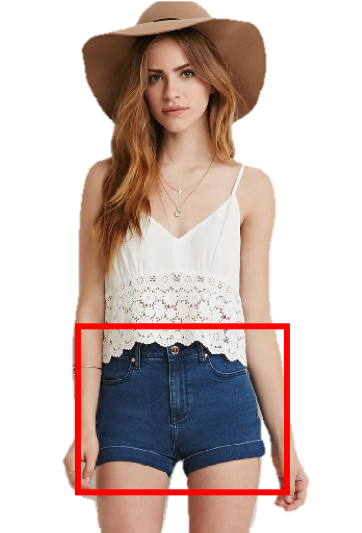}}}\hfill
\mpage{0.23}{\includegraphics[width=\linewidth, trim=65 0 65 0, clip]{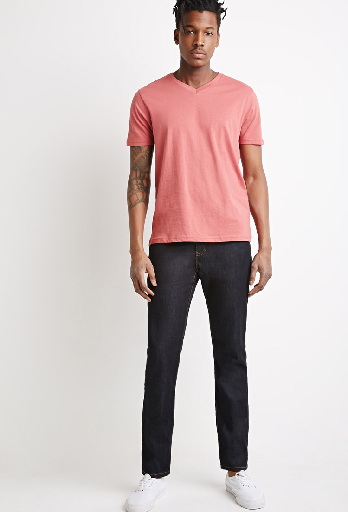}\llap{\includegraphics[height=1.75cm]{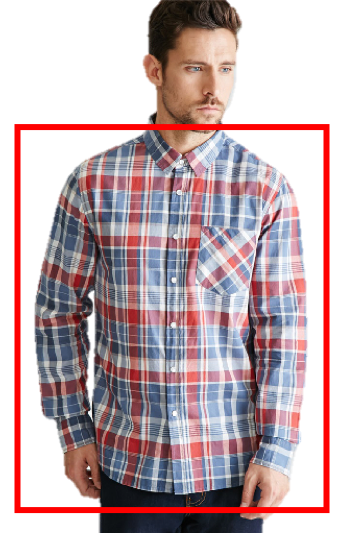}}}\hfill
\mpage{0.23}{\includegraphics[width=\linewidth, trim=65 0 65 0, clip]{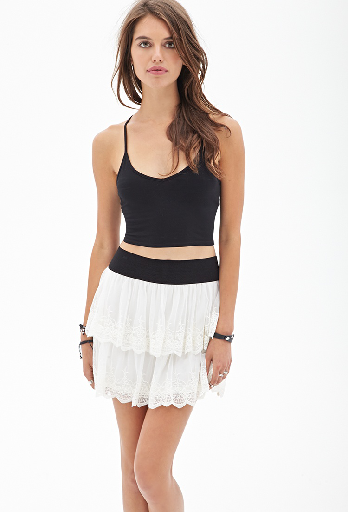}\llap{\includegraphics[height=1.75cm]{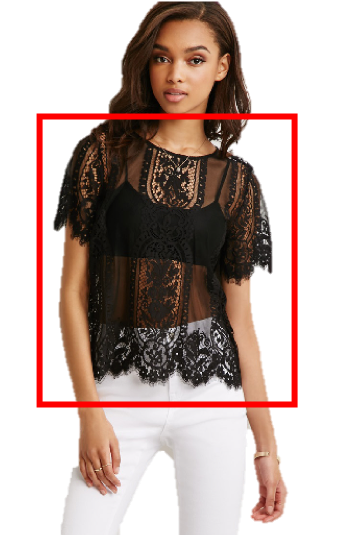}}}\\

\vspace{-.75mm}

\mpage{0.03}{\raisebox{0pt}{\rotatebox{90}{\small{Our results}}}}  \hfill
\hspace{-3mm}
\mpage{0.23}{\includegraphics[width=\linewidth, trim=65 0 65 0, clip]{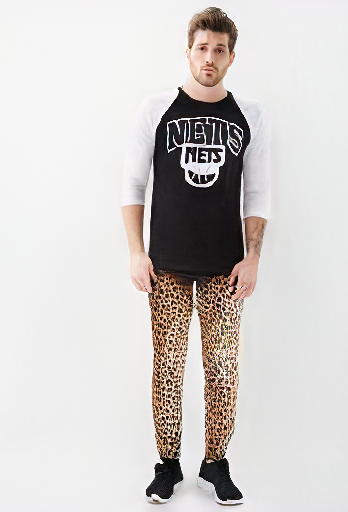}}\hfill
\mpage{0.23}{\includegraphics[width=\linewidth, trim=65 0 65 0, clip]{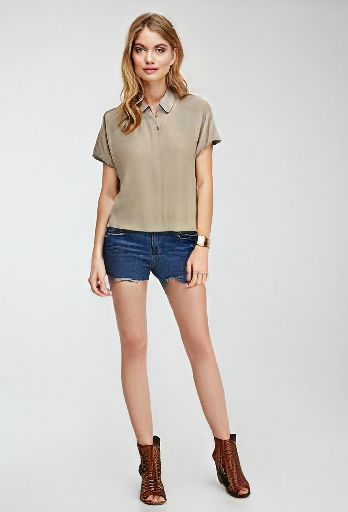}}\hfill
\mpage{0.23}{\includegraphics[width=\linewidth, trim=65 0 65 0, clip]{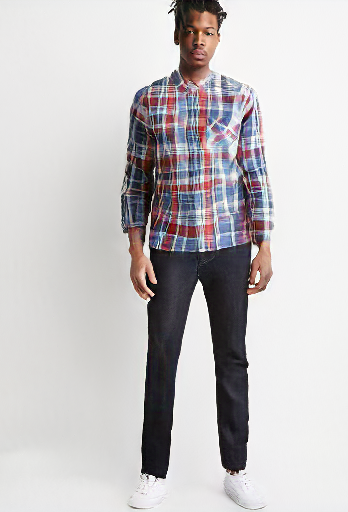}}\hfill
\mpage{0.23}{\includegraphics[width=\linewidth, trim=65 0 65 0, clip]{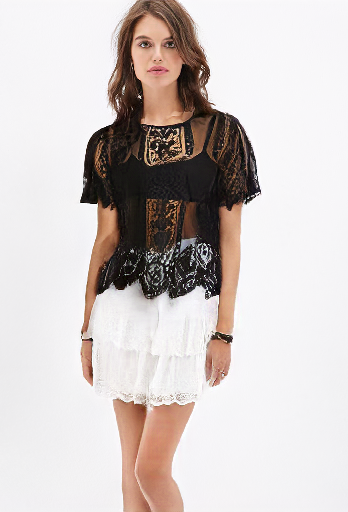}}\\

\vspace{-3mm}
\caption{\textbf{Garment transfer.}
We show examples of garment transfer for bottom (\emph{left}) and top (\emph{right}) garment sources. 
}
\label{fig:garment}
\end{figure}

Using the UV-space pre-computed mapping table, we can segment the UV-space into human body parts (Figure~\ref{fig:uv_parts}). 
We can then use this UV-space segmentation map to generate the target pose segmentation map using the target dense pose $P_{trg}$.
The target segmentation map allows us to combine partial features from multiple source images to perform garment transfer.
Figure~\ref{fig:garment} shows examples of bottom and top garment transfer.


\subsection{Limitations}
\begin{figure}[t]
\centering
\mpage{0.23}{\includegraphics[width=\linewidth, trim=50 0 50 0, clip]{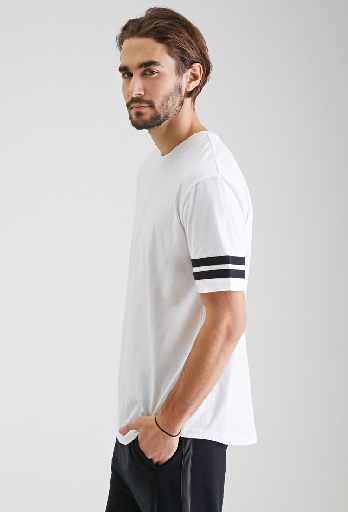}\llap{\includegraphics[height=1.7cm]{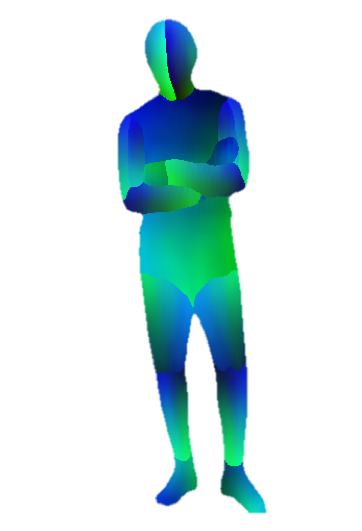}}}\hfill
\mpage{0.23}{\includegraphics[width=\linewidth, trim=50 0 50 0, clip]{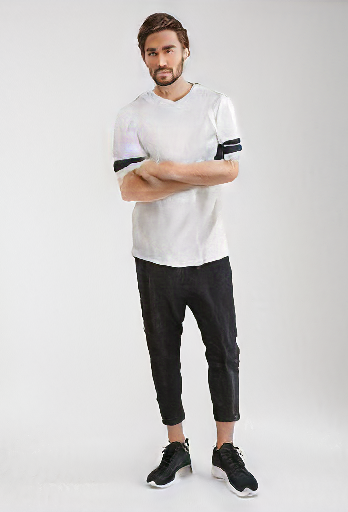}}\hfill
\mpage{0.23}{\includegraphics[width=\linewidth, trim=50 0 50 0, clip]{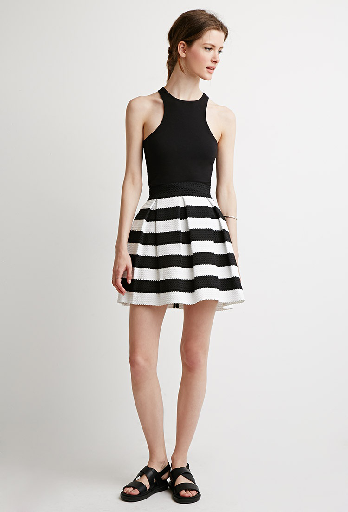}\llap{\includegraphics[height=1.7cm]{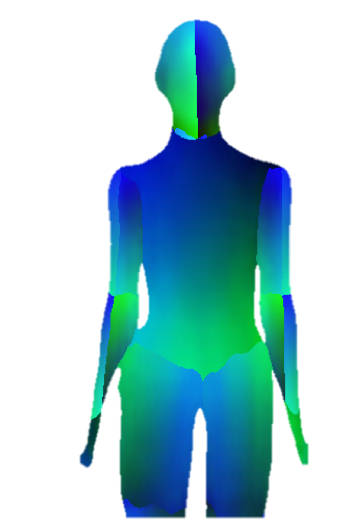}}}\hfill
\mpage{0.23}{\includegraphics[width=\linewidth, trim=50 0 50 0, clip]{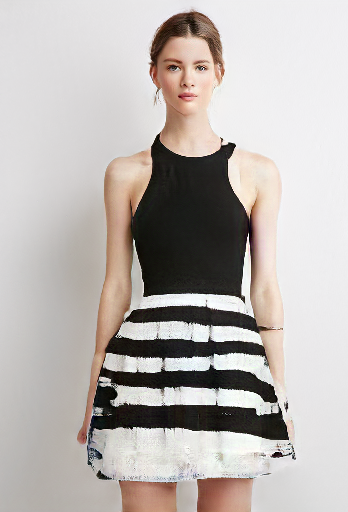}}\\
\vspace{-3mm}
\caption{\textbf{Failure cases.} Our method produces artifacts on the hands (\emph{left}) and the skirt (\emph{right}).
}

\label{fig:failure}
\end{figure}

\topic{Failure cases.}
Human reposing from a single image remains challenging. 
\figref{failure} shows two failure cases where our approach fails to synthesize realistic hands and clothing textures.
Hands are difficult to capture due to the coarse granularity of DensePose. Explicitly parsing hands could help establish more accurate correspondence between source/target pose (e.g., using Monocular total capture \cite{Xiang_2019_CVPR}).
Long-hairs and loose-fit clothes (e.g., skirts) are challenging because they are not captured by DensePose. We believe that incorporating human/garment parsing in our framework may help mitigate the artifacts. 

\begin{figure}[t]
\centering

\mpage{0.235}{\includegraphics[width=\linewidth, trim=0 0 0 0, clip]{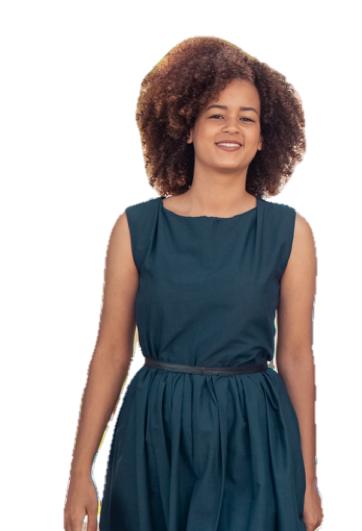}\llap{\includegraphics[height=1.1cm]{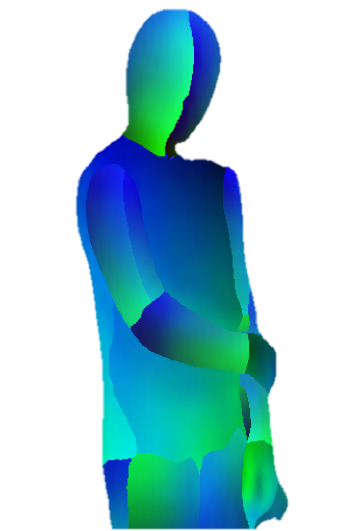}}}\hfill
\mpage{0.235}{\includegraphics[width=\linewidth, trim=0 0 0 0, clip]{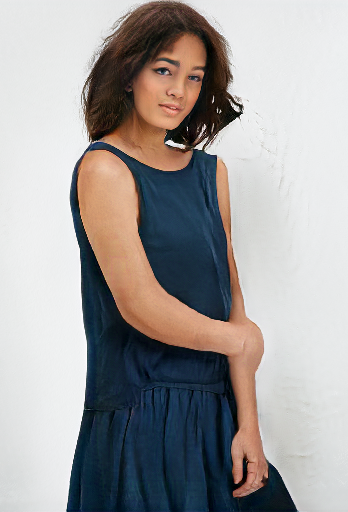}}\hfill
\mpage{0.235}{\includegraphics[width=\linewidth, trim=0 0 0 0, clip]{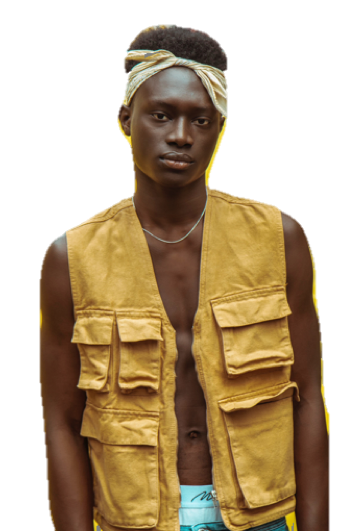}\llap{\includegraphics[height=1.1cm]{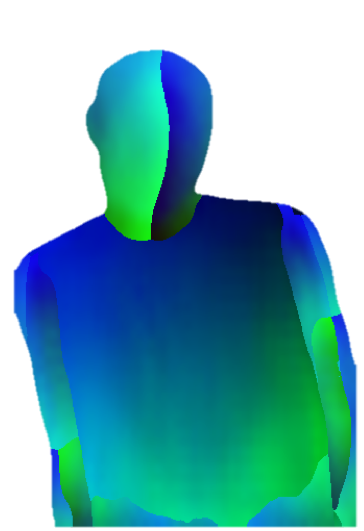}}}\hfill
\mpage{0.235}{\includegraphics[width=\linewidth, trim=0 0 0 0, clip]{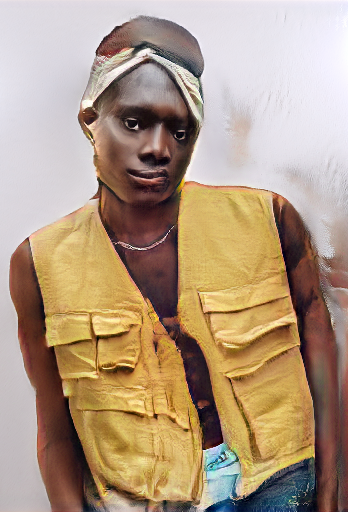}}\\
\vspace{-3mm}
\caption{\textbf{Diverse in the wild cases.} Our model inherits the biases of DeepFashion dataset and thus performs worse on unrepresented individuals. Our method cannot accurately synthesize curly hair (\emph{left}) and fails in reposing dark-skinned individuals (\emph{right}).
}
\vspace{-4mm}

\label{fig:diverse}
\end{figure}

\topic{Diversity}
DeepFashion dataset consists of mostly young fit models and very few dark-skinned individuals. 
Our trained model thus inherit the biases and perform worse on unrepresented individuals as shown in \figref{diverse}.
We believe that training and evaluating on diverse populations and appearance variations are important future directions. 

\section{Conclusions}
\label{sec:conclusions}

We have presented a simple yet effective approach for pose-guided image synthesis.
Our core technical novelties lie in 1) spatial modulation of a pose-conditioned StyleGAN generator and 2) a symmetry-guided inpainting network for completing correspondence field.
We demonstrate that our approach is capable of synthesizing \emph{photo-realistic} images in the desired target pose and \emph{preserving details} from the source image. 
We validate various design choices through an ablation study and show improved results when compared with the state-of-the-art human reposing algorithms.
Our controllable human image synthesis approach enables high-quality human pose transfer and garment transfer, providing a promising direction for rendering human images.

\bibliographystyle{ACM-Reference-Format}
\bibliography{main}

\end{document}